\documentclass[11pt]{article}
\usepackage[top=1in, bottom=1in, left=1in, right=1in]{geometry}

\usepackage[utf8]{inputenc} 
\usepackage[T1]{fontenc}    
\usepackage{enumitem}
\usepackage{algorithm}
\usepackage{algpseudocode}
\usepackage{tikz}
\usetikzlibrary{calc}
\usepackage{amssymb}
\usepackage{amsmath}
\usepackage{amsthm}
\usepackage{multirow}
\usepackage{booktabs} 
\usepackage{subfigure}
\usepackage{makecell}
\usepackage{arydshln}
\usepackage{float}

\newcommand\R{\mathbb R}

\newtheorem{theorem}{Theorem}[section]

\newtheorem{definition}{Definition}[section]
\newtheorem{lemma}[theorem]{Lemma}
\newtheorem{remark}[theorem]{Remark}

\newcommand{\eqlabel}[1]{\label{#1}\tag{#1}}

\newcommand{\norm}[1]{\ensuremath{\left\lVert #1 \right\rVert}}
\renewcommand{\norm}[1]{\lVert #1 \rVert}
\newcommand{\br}[1]{\left\{#1\right\}}           

\def\h_#1{\hat{#1}}
\def\wh_#1{\widehat{#1}}

\newcommand{\avg}{\textrm{avg}}
\renewcommand{\sim}{\textrm{sim}}
\newcommand{\avgSim}{\textrm{avgSim}}

\newcommand{\DS}{\textrm{DS}}

\newcommand{\DDS}{\textrm{DDS}}

\title{Implicit Diversity in Image Summarization}

\usepackage{authblk}

\author{L. Elisa Celis}
\author{Vijay Keswani}
\affil{Yale University}
\date{}

\begin{document}

\maketitle

\begin{abstract}
Studies have shown that the people depicted in image search results tend to be of majority groups with respect to socially salient attributes. This skew goes beyond that which already exists in the world - e.g., Kay et al. \cite{kay2015unequal} showed that although 28\% of CEOs in US are women, only 10\% of the top 100 results for CEO in Google Image Search are women. Most existing approaches to correct for this kind of bias assume that the images of people include socially salient attribute labels. However, such labels are often unknown. Further, using automated techniques to infer these labels may often not be possible within acceptable accuracy ranges, and may not be desirable due to the additional biases this process could incur.  

We develop a novel approach that takes as input a visibly diverse control set of images and uses this set to select a set of images of people in response to a query. The goal is to have a resulting set that is more visibly diverse in a manner that emulates the diversity depicted in the control set. Importantly, this approach does not require images to be labelled at any point; effectively, it gives a way to implicitly diversify the set of images selected. We provide two variants of our approach: the first is a modification of the MMR algorithm \cite{carbonell1998use} to incorporate the diversity scores, and second is a more efficient variant that does not consider within-list redundancy.

We evaluate these approaches empirically on two datasets 1) a new dataset containing top Google image results for 96 occupations, for which we evaluate gender and skin-tone diversity with respect to occupations and 2) the CelebA dataset \cite{liu2015faceattributes} for which we evaluate gender diversity with respect to facial features. Our approaches produce image sets that significantly improve the visible diversity of the results, compared to current Google search and other diverse image summarization algorithms, at a minimal cost to accuracy.

\end{abstract}

\newpage

\section{Introduction}

Services such as Google Image Search perform the task of image summarization; namely, responding to a query with an appropriate set of images. 
However, for queries related to people, such algorithms are often biased with respect to socially salient attributes\footnote{Sometimes called sensitive attributes or protected attributes in the fair machine learning literature.} of the data, such as the presented gender \cite{kay2015unequal, singh2019female} or skin tone \cite{buolamwini2018gender}. In essence, the summarization algorithms over-represent the majority demographics for a given query. Kay et al. \cite{kay2015unequal} show that such errors can reinforce the gender stereotypes associated with the query, underlining the need to correct such biases in image summarization results.
Furthermore, the use of demographically skewed results can be propagated and reinforced by other tools; 
e.g., state of the art image generation algorithms such as Generative Adversarial Networks (GANs), when trained on publicly available images of engineers, mostly generates images of white men wearing a hard hat \cite{engMisrepr}.
Clearly, it is crucial to ensure that the algorithms used for image summarization do not propagate or exacerbate societal biases.
To that end, our goal is to provide a simple and applicable image summarization algorithm which can ensure that the images  correspond to the query  yet are also visibly diverse.

\subsection{Our contributions}

In 2013-14, Kay et al. \cite{kay2015unequal} collected Google top 400 image results for each of the 96 occupations, and had 10\% of the images labeled by crowd workers according to presented gender.
They used this dataset to infer the gender-bias in the Google search results of occupations described above.
In the 6 years since then Google has continually updated its image analysis algorithms \cite{googlePhotos}. 
Hence, the first question we address is: \emph{does bias remain an issue in Google image search results?} 

Towards this, we consider the same 96 occupations, and collect the top 100 Google search results for each one in December of 2019.\footnote{http://bit.ly/2QVfM0K} 
We have these images labeled by crowd workers using Amazon Mechanical Turk (AMT) with respect to presented gender (coded as male, female, or other) and skin-tone (coded according to Fitzpatrick skin-tone scale).
This results in 60\% of images containing gender labels and 63\% of images containing skin-tone labels. While some improvements have been made with respect to gender (the \% of images of women in Google 2014 results is 37\% and in Google 2019 results it is 45\% ), we find that the fraction of gender anti-stereotypical images 
\footnote{{Anti-stereotypical images refer to set of images that do not correspond to the stereotype associated with the query. For example, gender anti-stereotypical images for a male-dominated occupation (determined using ground truth) would correspond to the set of images of women in the summary generated for that occupation.}} 
is still quite low (30\% in Google 2019 results and 22\% in Google 2014 results).
For skin-tone, 52\% of the images have a \textit{fair} skin-tone label (corresponding to Type 1-3 on the Fitzpatrick scale) and 10\% of the images have a \textit{dark} skin-tone label (corresponding to Type 4-6 on the Fitzpatrick scale).
Once again, the fraction of images of dark-skinned people in Google results is quite low.
Overall 57\% of the dataset has both a gender and skin-tone label; however, only 7\% of these are images of dark-skinned men and 3\% are images of dark-skinned women.
A final statistic that captures the lack of diversity in Google results is that 35 out of 96 occupations do not have any images of dark-skinned gender anti-stereotypical people in the top 100 results.
This assessment of Google images with respect to skin-tone was not possible for the original dataset of images from 2014, as no skin tone labels were present.

Given the extent and importance of this problem, the next question we address is: \emph{are there simple and efficient methods that correct for visible diversity across socially salient attributes in image search?}
When considering this question, we first note that, in general, images that contain people would not have their socially salient attributes explicitly labeled. 
Datasets are at scales where collecting explicit labels is infeasible, and while it may be possible to \emph{learn} these attributes in a pre-processing step, as we also observe this can lead to additional errors and biases \cite{scheuerman2019computers}.
Hence, we add a constraint to our main question: \emph{are there simple and efficient methods that correct for visible diversity across socially salient attributes in image search results which do not require or infer attribute labels?}
To the best of our knowledge, no methods with such a requirement exist for image summarization nor any other related problem in machine learning.
\footnote{
{
The goal of search algorithms is usually to return a ranking of images given an input query. 
While our approach can be extended to the case of ranking as well, in this paper, we will primarily focus on the task of fair retrieval, i.e., returning a fair \textit{summary} of images corresponding to an input query and ensuring that the top results are unbiased.
The reason for this simplification is to better analyze, highlight and mitigate the bias in the most visible results of image search, often characterized by images in the first or second page of the search results.
However, as discussed in Remark~\ref{rem:ranking}, our algorithms can be used to a rank images in a diverse manner as well.
}
}

To address this question, we design two algorithms: \textbf{MMR-balanced}, a modification of the well-know \textbf{MMR} algorithm \cite{carbonell1998use}, and \textbf{QS-balanced}, a simpler and more efficient algorithm inspired by the former. In both cases, the method takes a black-box image summarization algorithm and the dataset it works with, and overlays it with a post-processing step that attempts to diversify the results of the black-box algorithm.
To do so, our method takes as input a very small control set of visibly diverse images; the control set is  query-independent and should be carefully constructed to capture the kind of visible diversity desired in the output.\footnote{The size of the diversity control set can vary by application, but we show the efficacy of our method with small sets of size 8-25.}
On a high-level, the process is as follows (see also Figure~\ref{fig:model}): 
each image is given a query similarity score using the black-box algorithm, which corresponds to how well it represents the desired query.
The candidate images are also given a similarity score with respect to each image in the diversity control set using a given similarity scoring tool.
After adding the query similarity score to the diversity control scores, we rank the images by the combined score for each image in the control set and output the ones with the best scores.
As required, this results in a method which implicitly diversifies the image sets without having to infer or obtain socially salient attribute labels.

\begin{figure*}[t]
\centering     
\subfigure[Occupations dataset- Query: CEO]{
\includegraphics[width=0.515\linewidth]{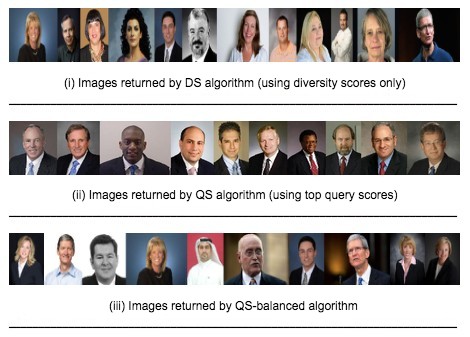}}
\subfigure[CelebA dataset- Query: Smiling]{
\includegraphics[width=0.445\linewidth]{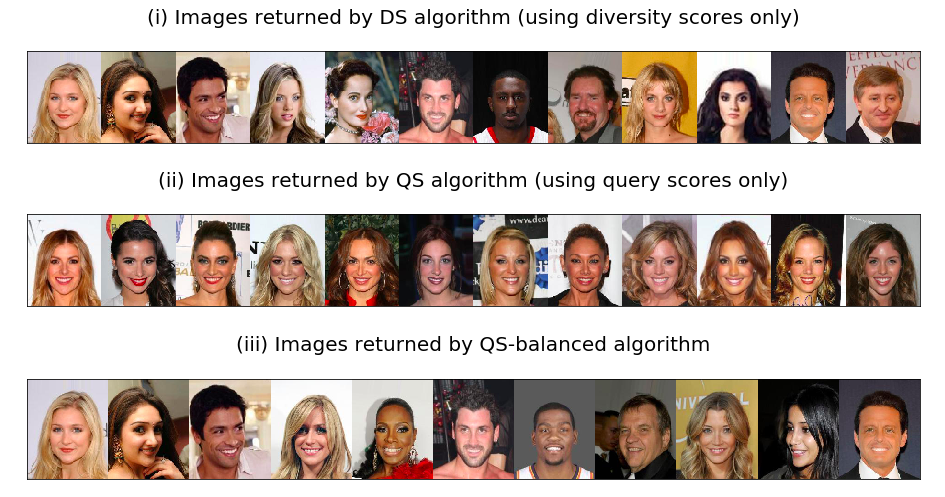}}
\caption{(a) Top images returned by \textbf{QS-balanced} for query ``CEO'' on Occupations dataset and (b) top images returned by \textbf{QS-balanced} for query ``smiling'' on CelebA dataset. The first row shows images returned by the algorithm using the diversity control matrix, the second row shows the images with most similarity to the query, the third rows shows images with best combined scores, i.e., minimum of $\br{\DS_{I_F}^q(I)}_{I \in S}$ for each $I_F$.
}
\label{fig:examples_ceo_doctor}
\end{figure*}

We evaluate the effectiveness of this approach on the new Occupations dataset we collect and the CelebA dataset.
The CelebA dataset contains more than 200,000 images of celebrities labeled with information about facial attributes of the person in the image.
For the Occupations dataset, the queries are the occupations while for the CelebA dataset, the queries are the facial attributes.

We compare the performance of our approaches on these datasets with other state-of-the-art algorithms and relevant baselines.
This includes summarization algorithms that reduce redundancy in the summary \cite{carbonell1998use}, or diversify across the feature space \cite{kulesza2012determinantal}, or use gender classification tools to compute explicit labels as a pre-processing step.
For the Occupations dataset, \textbf{QS-balanced} and \textbf{MMR-balanced} return more gender-balanced results than Google image search results (Section~\ref{sec:results_gender}) and baselines.
Specifically, the percent of gender anti-stereotypical images in the output of  \textbf{QS-balanced} and \textbf{MMR-balanced} is around 45\% on average across occupations, while for Google Image search, this number is approximately 30\%.
The baseline algorithms also have a relatively lower percent of gender anti-stereotypical images in their output (35\%-39\%), confirming observations made in prior work which state that diversifying across feature space or using pre-trained gender classification tools do not necessarily result in diversity with respect to socially salient attributes \cite{celis2016fair, scheuerman2019computers}.
Similarly, on the CelebA dataset, our algorithms return much more gender-balanced results, compared to the results using just query similarity or other algorithms.
In this case, the average fraction of gender anti-stereotypical images in the output of  \textbf{QS-balanced} is 0.23, while using just query similarity, this number is 0.08.
For example, for gender-neutral facial attributes, such as 'smiling', the 50 images obtained using top query scores are images of women while \textbf{QS-balanced} returns an image set with a 32\% men and no loss in accuracy.
On the Occupations dataset, we also show that \textbf{QS-balanced} and \textbf{MMR-balanced} increase the diversity across skin-tone as well as diversity across the intersection of skin-tone and gender.\footnote{The CelebA dataset does not contain race or skin tone labels, hence we cannot evaluate its performance with respect to these attributes.}
The average fraction of images of dark-skinned people in the output \textbf{QS-balanced} is 0.17 while for Google results, the average fraction is 0.16. 
However, the standard deviation is higher for Google results (0.09 vs 0.05), implying that the results are relatively more unbalanced for Google.
In terms of intersectional diversity, Table~\ref{tbl:int_div} shows that the results from \textbf{QS-balanced} algorithm are gender-balanced across skin-tone, unlike Google results.
The average fraction of images of dark-skinned gender anti-stereotypical images in the output of \textbf{QS-balanced} is 0.08 while for Google, this number if 0.05.

\begin{table}[t]
\centering
\small
\begin{tabular}{ |c |c |c |}
\hline
 \multicolumn{3} {|c|} {\bfseries Our Algorithm} \\ 
 \hline
 & \makecell{\% gender\\ stereotypical} & \makecell{\% gender\\ anti-stereotypical}  \\
\hline
Fair skin & 0.46 (0.14)  &  0.37 (0.14)\\
Dark skin & 0.09 (0.05) & 0.08 (0.05)\\
\hline
\end{tabular}
\quad
\begin{tabular}{ |c |c |c |}
\hline 
 \multicolumn{3} {|c|} {\bfseries Google Images} \\ 
 \hline
 & \makecell{\% gender\\ stereotypical} & \makecell{\% gender \\anti-stereotypical}  \\
\hline
Fair skin & 0.60 (0.20) &  0.24 (0.21) \\
Dark skin & 0.11 (0.08) & 0.05 (0.07)\\
\hline
\end{tabular}\\
\caption{{Comparison of intersectional diversity of top 50 \textbf{QS-balanced} images and Google images.}
The number represents the average fraction of images satisfying the corresponding attribute, with standard deviation in the brackets.%
Google images seems to have a larger fraction of stereotypical images, with respect to both gender and skin-tone.
In comparison, \textbf{QS-balanced} returns images that are relatively more balanced; for both skin-tones, the fraction of men and women in the output is almost balanced.
Intersectional diversity comparison with other baselines is presented in Table~\ref{tbl:int_div_all}.}
\label{tbl:int_div}
\end{table}

However, the increase in diversity with respect to skin-tone is limited, perhaps due to the lack of skin-tone diversity in the dataset itself. We show that we can improve on these numbers by more aggressively weighting the diversity score (computed with respect to diversity control set), this comes at an increased cost to accuracy. 
Importantly, our focus is on visible diversity, i.e., presented gender and skin color.
We make this choice as true labels are often not only unknown but also irrelevant -- e.g., a set of images of male-presenting CEOs is not sufficiently diverse to combat the problems mentioned above, regardless of the true gender identity of the people captured in the images.

The following sections are organized as follows: after briefly reviewing related work in the field of diverse image summarization, we start with a description of the setting of summarization, followed by the details of our suggested algorithms in Section~\ref{sec:model}.
We next present the Occupations dataset and assess the gender and skin-tone diversity of the dataset in detail in Section~\ref{sec:dataset}.
Following this, we state at the results of the empirical analysis of our algorithm on the Occupations and CelebA dataset (Section~\ref{sec:experiments}).
Finally, we discuss the implications and inferences from our results and address the limitations of our methods and ways to improve it in future work (Section~\ref{sec:limitations}).

\subsection{Background and related work}
{
To assess the importance of addressing bias in summarization results, we first look at prior work on the social impact of stereotypes, as well as, related work in the field of fair machine learning.

\subsubsection{Impact of stereotypes}
The study of cultivation and the impact of stereotypes has drawn serious interest in the age of digital media \cite{potter1994cultivation, o2016weapons}, primarily due to the increased ease of information access and the possibility of stereotype propagation via sources like images on social media, search results, etc.
To define briefly, stereotyping is the process of inferring common characteristics of individuals of a group. When used accurately, stereotypes associated with a group are helpful in deducing information about individuals from the group in the absence of additional information \cite{mcgarty2002social, bodenhausen1985effects} and also function as tools to characterize group action \cite{haslam2002personal, cadinu2013comparing, tajfel2001social}.
However, inaccurate or exaggerated stereotypes can be quite harmful and can inadvertently lead to biases against the individuals from the stereotyped group. 
Prior studies have shown that association of a negative stereotype with a group for a given task can affect the performance of the stereotyped individuals on the task \cite{spencer1999stereotype, word1974nonverbal}; using the performance on such a task for any kind of future decision-making will lead to the propagation of such stereotypes and bias the results against one group.
Furthermore, inaccurate stereotypes also lead to an incorrect perception of reality, especially with respect to sub-population demographics \cite{shrum1995assessing, gerbner1986living, kay2015unequal}.
For example, stereotypical images of Black women as matriarchs or mammies, that are further disseminated via digital media, can lead to such a negative stereotype seem natural \cite{collins2002black, harris1982mammies}.
Given the existence of such negative social stereotypes and the possibility of their propagation via images, it is important to question the presence of stereotypes in image summary representations and explore methods to prevent the exacerbation of social biases through image search results of people.

\subsubsection{Bias in existing image datasets and models}
The effect of negative stereotypes and the resulting biases have been carefully explored in television media in the form of \textit{cultivation theory} \cite{shrum1995assessing,gerbner1986living}, particularly with respect to the portrayal of women, racial and ethnic minorities.
Online media has only recently been subjected to similar scrutiny and multiple studies have highlighted the presence of such biases in existing summarization tools and benchmark image datasets.

As discussed before, the study by Kay et al. \cite{kay2015unequal} explored the effects of bias in Google image search results of occupations on the perception of people of that occupation.
Follow-up studies by Pew Research Center \cite{pewGender} and Singh et al. \cite{singh2019female} also found evidence of similar gender-bias in Google image search results; \cite{pewGender} further observed that, for many occupations, images of women tend to appear lower than the images of men in search results.
Biased representation of minorities has also been observed in other computer vision applications.
Buolamwini and Gebru \cite{buolamwini2018gender} found that popular facial analysis tools from IBM, Microsoft and Face++ have a significantly larger error rate for dark-skinned women than other groups.
This study led to a subsequent improvement in the accuracy of these tools with respect to images of minorities \cite{ibmResponse} and it highlights the importance of constant audit of existing models, as well as, the need for alternative strategies to develop unbiased models, since even improvements to existing facial analysis tools do not achieve desired diversity in their results.
A case in point is the study by Scheuerman, Paul and Brubaker \cite{scheuerman2019computers} which showed that commercial facial analysis tools do not perform well for transgender individuals and are unable to infer non-binary gender.

Even existing datasets, collected from real-world settings, can encode unwarranted biases that can occur from the data collection process.
Van Miltenburg \cite{van2016stereotyping} provided evidence of stereotype-bias in a popular dataset of Flickr images annotated with crowdsourced descriptions.
The study by Zhao et al. \cite{zhao2017men} found that datasets used for visual recognition tasks have significant gender bias. 

\noindent
\subsubsection{Downstream propagation of biases}
As mentioned earlier, inaccurate representation of demographic groups can lead to biases against these groups, either in the form of incorrect perceptions about the group \cite{kay2015unequal, collins2002black, harris1982mammies} or in the form of bias in decision-making process based on the inaccurate representations. \cite{potter1994cultivation, crawley2014gender, ramaci2017gender, bear2017performance, kanze2018we}.
If a machine learning model is trained using an imbalanced or misrepresentative dataset, the biases in the dataset can edge into the output of the model as well.
For example, Datta et al. \cite{datta2015automated} showed that men are more likely to be shown Google ads for high-paying jobs than women, a result of training the targeting model on gender-biased data.
Similarly, Caliskan et al. \cite{caliskan2017semantics} found that word associations learned from existing texts encode historical biases, such as gender stereotypes for occupations.
Image generation algorithms, such as GANs \cite{karras2018style}, when trained on Google Images of people from certain common occupations, mostly generate stereotypical images \cite{engMisrepr}.
With any additional intervention, unconstrained models, including summarization algorithms,  are bound to reflect the biases of the dataset they operate upon.
Hence, to prevent the propagation of bias due to imbalanced image summaries, it is important to develop summarization algorithms that ensure that the generated summaries are unbiased even when using biased datasets.

\noindent
\subsubsection{Algorithms for image summarization}
The rising popularity of social networks and image hosting websites has led to a growing interest in the task of image summarization.
The primary goal of any image summarization algorithm is to appropriately condense a given set of images into a small representative set.
This task can be divided into two parts: (a) scoring images based on their importance, and (b) ensuring that the summary represents all the relevant images.

Traditional image summarization algorithms to score images on their importance have focused on using visual features, such as color or texture, to compare and rank images \cite{hafner1995efficient, yang2011effective}.
Recently, even pretrained neural networks have been used for image feature extraction \cite{shin2016deep}, which is then used to score images based on their \textit{centrality} in the dataset.
In case of query-based summarization, determining importance of an image includes determining whether the image is relevant to the query.
To find query-relevant images, search services like Google use metadata from the parent websites of images to associate keywords with them, thus simplifying the task significantly \cite{yu2014joint}.
However, for the datasets we analyze, metadata or keywords for images are not available; correspondingly we need to use retrieval algorithms that use image features only.
If the queries come from a pre-determined set, then supervised approaches for image classification can also be used for summarization \cite{rozsa2019facial, rawat2017deep, zhang2016gender, ehrlich2016facial}.
For example, if the queries correspond to facial features, then scores from state-of-the-art convolutional neural networks pre-trained on large image datasets with annotated facial features \cite{liu2015faceattributes} can be employed for retrieving relevant images.
We will show the efficacy of such an approach in Section~\ref{sec:experiments} for the CelebA dataset.
In the absence of pre-trained classification models and metadata information, one has to adopt unsupervised approaches to determine query relevance of images.
Given a query image, an unsupervised approach suggested by \cite{wiggers2019deep} uses pre-trained models \cite{imagenet} to find images similar to the query image; they show that this unsupervised approach is comparable to state-of-the-art algorithms for the task of \textit{pattern spotting}.
We will use this approach for query-based summarization for the Occupations dataset.

Secondly, to ensure that the summary is representative of all relevant images, most prior work have used the idea of \textit{non-redundancy} \cite{carbonell1998use, radlinski2009redundancy, sanderson2009else, clarke2008novelty, lin2011class}.
Once the images have been scored on their relevance, algorithms such as MMR \cite{carbonell1998use}, greedily select images that are not very similar to the images already present in the summary.
Other efficient methods to ensure non-redundancy in the summary include the use of determinantal point processes \cite{kulesza2012determinantal} and submodular maximization models \cite{tschiatschek2014learning}.
These models have also been used explicitly for the task of efficiently summarizing images of people \cite{sinha2011extractive}.
However, reducing redundancy in the output set does not always correspond to diversity with respect to the desired features, such as gender, race, etc., as demonstrated by Celis et al. \cite{celis2016fair}.
Our evaluations using redundancy-reducing algorithms also lead to this conclusion.
We empirically compare our algorithm to such non-redundancy-based approaches in Section~\ref{sec:experiments} and discuss them further in Section~\ref{sec:limitations}.
}

\noindent
\subsubsection{Prior work on unbiased image summarization}
Current approaches to debias summarization algorithms often assume the existence of socially salient attribute labels for data-points.
Lin et al. \cite{lin2011class} suggest a scoring function over subsets of elements that rewards subsets that have images from different partitions.
For example, Celis et al. \cite{celis2018fair} formulate the summarization problem as sampling from a Determinantal Point Process and use partition constraints on the support to ensure fairness.
However, setting up the partition constraints or evaluating scores requires the knowledge of the partitions and correspondingly the socially salient attributes for all data-points.
Similarly, fair classification algorithms, such as \cite{celis2018classification, corbett2017algorithmic, dwork2012fairness, hardt2016equality, kamiran2012data, zafar2015fairness, zhang2018mitigating} use the gender labels during the training process.
Even for language-based image recognition tasks, \cite{zhao2017men} suggest constraints-based modifications of existing models to ensure fairness of these models, but the constraints are based on the knowledge of the gender labels.
Unlike these approaches, our paper aims to ensure diversity in settings where socially salient attribute labels are not available.

\begin{figure*}
\centering
\includegraphics[width=\linewidth]{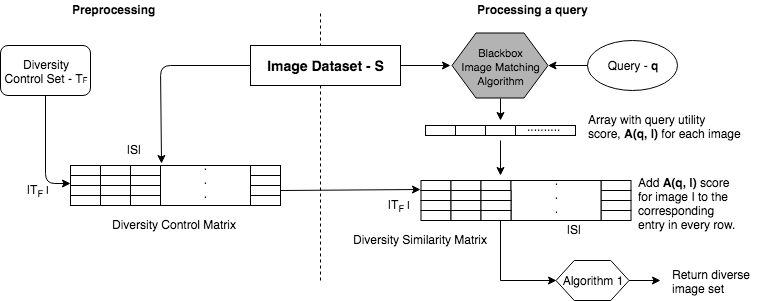}
\caption{A simple post-processing approach for ensuring diversity in image search. A small ``diversity control set'' of images is taken as input, and (relevant) images are assigned a similarity score with each image in the control set to create a diversity control matrix. These scores are combined with the query scores provided in a black-box manner using an existing image search approach. A summarization algorithm then selects the final images using this combined score. See Algorithm~\ref{algo:main} for details.}
\label{fig:model}
\end{figure*}

\section{Our model and algorithms} \label{sec:model}

In this section, we describe our approach to ensuring that the image summarization process returns visibly diverse images.
Given a query from the user, we start with the goal of choosing images that correspond to the query and then incorporate an additional novel diversity check (using a diversity control set provided by the user) into the model.
Let $S$ denote the large corpus of images.

\vspace{0.1in}
\noindent
\textbf{Query Score: }Suppose we are given a query $q$ and a black-box algorithm $A$ that takes the query and the dataset $S$ as input and for each image, returns a query similarity score; the score represents how well the image corresponds to the query.
The smaller the score $A(q, I)$, for a query $q$ and image $I$, the better the image corresponds to the query.
Since our framework is meant to extend an existing image retrieval model, we can assume that such a score can be efficiently computed for each query and image pair.

\vspace{0.1in}
\noindent
\textbf{Image Similarity Score:} Suppose that we also have a generic image similarity function $\sim$, which takes as input a pair of images, $I_1, I_2$, and calculates a score of similarity of the two images, $\sim (I_1, I_2)$. 
For the sake of consistency, here again we will assume that the smaller the score, the more similar are the images.

While the framework we propose is independent of the query matching algorithm or the image similarity function, we will present a concrete example of such algorithms and functions in a later section. We first see how we can use this score to rank our dataset.

\vspace{0.1in}
\noindent
\textbf{Diversity using a control set: }
A ranking/summary with respect to the scores returned by $A$ is unlikely to be visibly diverse without further intervention in most cases, as shown by prior studies \cite{kay2015unequal}.
To ensure visible diversity in the results, we use a diversity control set $T_F$ and a clustering approach.
The diversity control set $T_F$ is a \textit{small} set of visibly diverse images and will be used to enforce the diversity in the output; for example, if the summary is required to be gender-diverse, then the diversity control set will have equal number of images of men and women.

For each control image $I_F \in T_F$, using $\sim(\cdot, \cdot)$ as the distance metric, we can learn the cluster of images around $I_F$, by sorting $\br{\sim(I_F, I)}_{I \in S}$ for each $I_F \in T_F$.
In other words, we can associate each image $I \in S$ to an image in the control set to which $I$ is most similar.

\begin{algorithm}[bt]
   \caption{Ensure diversity in image summarization: QS-balanced}
\begin{algorithmic}[1]
   \State {\bfseries Input:} Dataset $S$, query $q$, query matching algorithm $A$, similarity function $\sim$, diversity control set $T_F$, parameter $\alpha \in [0,1]$, number of elements to be returned $M < |S|$
   \State For each $I \in S$, use black-box query algorithm to obtain score $A(q,I)$
   \State For each $I \in S$ and $I_F \in T_F$, calculate $\sim(I,I_F)$
   \State For each $I \in S$ and $I_F \in T_F$, calculate $\DS_{I_F}^q(I)$ using $\sim(I,I_F)$ and $A(q,I)$
   \State $T_R \gets \emptyset$
   \While{$|T_R| < M$}
   		\State $T \gets \emptyset$
   		\ForAll{$I_F \in T_F$} \Comment{Find images for each $I_F$} 
   			\State $J \gets \arg \min_{I \in S} \DS_{I_F}^q(I) $
   			\If{$J \notin T_R$} \Comment{Checking duplicates} 
	   			\State $T \gets T \cup \br{J}$
	   			\State score$(J) \gets  \DS_{I_F}^q(J)$  \Comment{Scores used for tie-breaks} 	
   				\State $\DS_{I_F}^q(J) \gets \infty$
	   		\EndIf
   		\EndFor
   		\If{$|T_R \cup T| \leq M$}  \Comment{If all of $T$ can be added} 
   			\State $T_R \gets T_R \cup T$
   		\Else \Comment{Tie-break when $M$ is not a multiple of $T_F$} 
   			\State $M' \gets M - |T_R|$ 
   			\State $T' \gets M'$ images from $T$ with lowest values of score$(J)$
   			\State $T_R \gets T_R \cup T'$   			
   		\EndIf
   \EndWhile
   \State  return $T_R$
\end{algorithmic}
   \label{algo:main}
\end{algorithm}

\vspace{0.1in}
\noindent
\textbf{Using diversity control sets with existing redundancy-reducing algorithms: }
To ensure we take into account both the query score from the black-box $A$ and the diversity with respect to the control set $T_F$, we have to combine the scores $A(q, \cdot)$ and $\sim(I_F, \cdot)$.
As mentioned earlier, a popular approach to combining query similarity and diversity is to diversify across the entire feature space, i.e., reduce the \textit{redundancy} of the chosen summary.
Using \textit{maximum marginal relevance} score is one of the many simple and efficient greedy selection procedures for this task \cite{carbonell1998use}.
Maximum marginal relevance (MMR) score of an image is a combination of the query similarity score of that image and its dis-similarity to the already chosen images; at every step, the image that optimizes this score is added to the set.
However, reducing redundancy does not necessarily lead to diversification across the desired attributes, such as gender \cite{celis2016fair}.
An obvious question in this respect is whether the diversity control set score can be incorporated with a non-redundancy approach to achieve diversity across gender, race, etc.

To that end, we present the \textbf{MMR-balanced} algorithm. 
Starting with an empty set $R$, the algorithm adds one image to the subset $R$ in each iteration.
The chosen image is the one that minimizes the score
\[ (1 - \alpha -\beta) \cdot A(q, I)  - \beta \cdot \min_{J \in R} \sim(I,J) + \alpha \cdot \min_{I_F \in T_F} \sim(I,I_F),\]
where $\alpha, \beta \in [0,1]$.
The first part of the above expression captures query relevance while the second part penalizes an image according to similarity to existing images in the summary $R$.
These two terms together constitute the \textit{maximum marginal relevance} score \cite{carbonell1998use}.
The third term in the above expression now acts as a deterrent to choosing multiple images corresponding to the same control set image $I_F$ (unless there are almost equal number of images corresponding to each $I_F$ in $R$).
The complete algorithm is formally presented in Algorithm~\ref{algo:mmr_bal}.
We will set $\alpha = \beta = 0.33$ for \textbf{MMR-balanced} in the following sections and empirical analysis.
We also analyze this expression theoretically in Section~\ref{sec:discussion}.

A drawback of \textbf{MMR-balanced} is the time complexity.
In particular, checking redundancy with existing images at every step is cumbersome and often necessary, if the dataset is diverse enough.
Furthermore, dropping the redundancy check should not affect the diversity with respect to socially salient attributes, since we have the diversity control term for that purpose.
This leads us to a more efficient algorithm.

\begin{algorithm}[bt]
   \caption{MMR-balanced }
\begin{algorithmic}[1]
   \State {\bfseries Input:} Dataset $S$, query $q$, query matching algorithm $A$, similarity function $\sim$, diversity control set $T_F$, parameter $\alpha, \beta \in [0,1]$, number of elements to be returned $M < |S|$
   \State $T_R \gets \emptyset$
   \While{$|T_R| < M$}
   		\ForAll{$I \in S \setminus T_R$} 
   			\State $\text{redundancy-score} \gets \min_{J \in T_R} \sim(I,J)$
   			\State $\text{diversity-score} \gets \min_{I_F \in T_F} \sim(I,I_F)$
			\State $\text{score}(I) \gets (1 - \alpha -\beta) \cdot A(q, I)  - \beta \cdot \text{redundancy-score}+ \alpha \cdot \text{diversity-score}$   			
   		\EndFor
   		\State $T_R \gets T_R \cup \arg\min_I \text{score}(I)$
   \EndWhile
   \State  return $T_R$
\end{algorithmic}
   \label{algo:mmr_bal}
\end{algorithm}

\vspace{0.1in}
\noindent
\textbf{QS-balanced: }
Given a tradeoff parameter $\alpha \in [0,1]$ and a query $q$, for each $I_F \in T_F$ let $\DS_{I_F}^q : S \rightarrow \R$ denote the following score function:
$$\DS_{I_F}^q(I) =         (1 - \alpha) \cdot A(q, I) +  \alpha \cdot \sim(I_F, I).$$
The score $\DS_{I_F}^q(I)$ corresponds to a combination of similarity with $I_F$ and similarity with a query $q$.
Finally, for each $I_F \in T_F$, we sort the set $\br{\DS_{I_F}^q(I)}_{I \in S}$ and return equal number of images with the lowest scores from each set, checking for duplicates at every step. 
The ties are broken by choosing the image with the better query score.
This gives us our final set of visibly diverse images. 
Algorithm~\ref{algo:main} formally summarizes this approach.
For $\alpha = 0.5$ and given a diversity control set, we will call this algorithm \textbf{QS-balanced}.
We will also refer to the algorithm using only diversity scores, i.e., $\alpha = 1$, as \textbf{DS} and the algorithm using only query scores, i.e., $\alpha = 0$, as \textbf{QS} in the following sections.

\vspace{0.1in}
\noindent
\textbf{Time complexity of the QS-balanced: }
Without making any assumption on the blackbox query relevance algorithm $A$, we can upper bound the additional time to ensure diversity using the control set.
The additional overhead in time complexity is $|T_F| \cdot \log(|S|) \cdot \mathcal{T}$, where $\mathcal{T}$ is the time taken to compute the similarity score for any given pair of elements.
This factor is due to the time taken to construct and sort the rows of the diversity-similarity matrix.
The time complexity also depends linearly on the size of the control set and hence the size of the control set should be much smaller than the size of the dataset.
Note that \textbf{MMR-balanced} is $O(M)$ times slower than \textbf{QS-balanced}, where $M$ is the size of the summary.

\vspace{0.1in}
\noindent
\textbf{Model Properties: }
An important property that many diverse summarization algorithms (including \textbf{MMR}) share is the \textit{diminishing returns property}  \cite{carbonell1998use, lin2011class, tschiatschek2014learning}.
To state briefly, a function, defined over the subsets of a domain, satisfies the\textit{diminishing returns property} if the change in function value on adding an element to smaller set is relatively larger.
Such set functions are also called \textit{submodular functions}.
Due to diminishing returns property, simple greedy algorithms can be used to approximately and efficiently optimize these functions, making them ideal for summarization over large datasets.

We can directly show that score computed at each step of \textbf{MMR-balanced} satisfies the diminishing returns property (simple extension of proof for \textbf{MMR}).
Even \textbf{QS-balanced}, if represented as an iterative process, can be shown to satisfy this property, implying that these algorithms share the mathematical features of common diverse summarization algorithms and that fast and greedy approaches do lead to approximately good solutions.
We formalize these statements in Section~\ref{sec:discussion} and provide mathematical proofs of submodularity of these functions.

\begin{remark}[Ranking] \label{rem:ranking}
{
Search algorithms usually return a ranking of images in the dataset and ranking models also suffer from the same kind of biases studied in the case of summarization \cite{pewGender, celis2017ranking}.
While ranking a set of images can be considered an extension of the summarization problem,
we primarily focus on summarization to highlight and mitigate bias in the most visible results of image search.
However, given the similarity between these problems, an obvious question is whether our approach can be used to provide a \textit{fair ranking} of the images.
Indeed, both \textbf{QS-balanced} and \textbf{MMR-balanced} can be used to rank images as well. 
Both algorithms inherently compute a score for each image which captures both the query similarity and diversity with respect to the control set (see Section~\ref{sec:discussion} for more details).}
While the \textbf{QS-balanced} is for diverse image summarization, with slight modification the algorithm can also be used to rank the images in the dataset according to the score $\DS_{I_F}^q(I)$.
We can construct a $|S| \times |T_F|$ sized matrix (as shown in Figure~\ref{fig:model}) with the entry corresponding to $I, I_F$ storing the score $\DS_{I_F}^q(I)$.
Next we first sort each row of this matrix according to the stored score and then sort each column.
Finally we can assign a ranking, starting with the image corresponding to the first entry of the matrix and moving along the first column. 
Once the first column has been ranked, we move to the second column and so on, checking for duplicates at each step.
\end{remark}

\section{Datasets} \label{sec:dataset}

\subsection{Occupations dataset}
We compile and analyze a new dataset of images for different occupations.
The dataset is composed of the top 100 Google Image Search results \footnote{The images were collected in December 2019.} for 96 different occupations.
This dataset is an updated version of the one compiled by Kay, Matuszek and Munson \cite{kay2015unequal}, which contained Google image results from 2013 \footnote{https://github.com/mjskay/gender-in-image-search}. 

Since occupations are often associated with gender or race-stereotypes, empirical analysis with respect to these search terms will help better evaluate the imbalance in existing search and summarization algorithms.
To compare the composition of the dataset with the ground truth of the fraction of minorities working in the occupation, we use the census data of the fraction of women and Black people working in each occupation from the Bureau of Labor and Statistics \cite{bls}.
The census data shows that Black people are the racial minority in each of the considered occupation (relative of White people). On the other hand, 52 out of 96 occupations have a larger fraction of men employed and the rest have a larger fraction of women employed.
In our analysis, we will often compute the fraction of gender anti-stereotypical images for different occupation, i.e., if an occupation is male-dominated, we take into account the fraction of women and if an occupation is female-dominated, we take into account the fraction of men in the output set.

We use Amazon Mechanical Turk to label the gender and Fitzpatrick skin-tone of the primary person in the images.
To obtain labels, we designed a survey asking participants to label the gender and skin-tone of the primary person in the images.
Each survey had around 50 images and the surveys were limited to participants in US.
Since some of the images had multiple primary persons or people whose features were hidden or cartoon images, ``Not applicable'' and ``Cannot determine'' were also provided for each question.
For each image, we collect 3 responses and assign the majority label to the image.

{
We use standard inter-rater reliability measurements to quantify the extent of consensus amongst different participants of the survey. 
Overall there were around 620 survey participants and each participant only labels a small subset of images (50).
We compute the Cohen's $\kappa$-coefficient \cite{cohen1960coefficient} for all pairs of participants with more than 5 common images in their surveys. 
\footnote{Similar techniques to evaluate interrater agreement in the setting of multiple participants rating a subset of elements has been considered in other prior work as well \cite{levy2018towards, mulki2019hsab}.} 
The resulting mean $\kappa$-coefficient across the pairs is 0.58 (median is 0.62).
Based on existing heuristic guidelines and interpretations of these coefficients \cite{landis1977measurement}, these results imply that, on an average, there is a \textit{moderate} level of agreement between survey participants.
}

An analysis of this dataset revealed similar diversity results as the analysis by Kay et al. \cite{kay2015unequal} of Google images from 2013.
However, while their analysis was limited to gender, we are also able to assess the skin-tone diversity of the results.
{Furthermore, unlike Kay et al., who mainly report the fraction of images of women in top results, we focus on measuring the fraction of gender anti-stereotypical images in top images.
This is because our primary goal is to provide balanced summariries and present anti-stereotypical images to effectively counter gender-stereotypes \cite{finnegan2015counter}. Measuring the fraction of anti-stereotypical images better quantifies the stereotype-exaggeration in current results, compared to fraction of images of women.}

\begin{figure*}[t]
\centering     
\subfigure[Fraction of images of women in Google results]{
\includegraphics[width=0.48\linewidth]{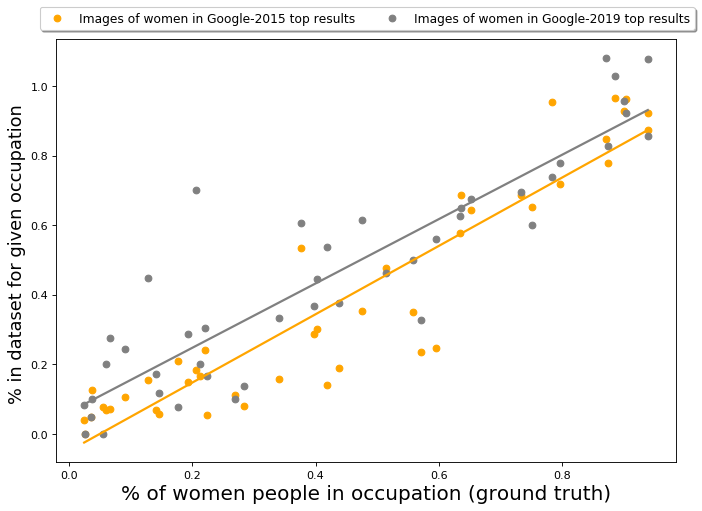}}
\subfigure[Fraction of images of dark-skinned people and dark-skinned women in Google results]{
\includegraphics[width=0.48\linewidth]{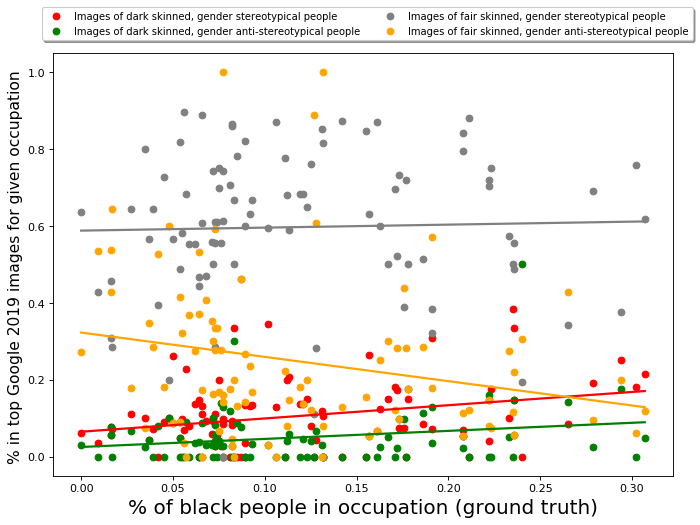}}
\caption{The plots show the fraction of images of women, dark-skinned people and their intersection in the top 100 results of Google Image Search.
(a) For gender, we also provide the comparison with Google results from 2013 \cite{kay2015unequal}.
While the fraction of women in the top Google results seems to have increased, the fraction of gender stereotypical images is still high (~0.7 on an average).
(b) Majority of the top Google images for every occupation corresponds to gender-stereotypical fair-skinned people, independent on the ground truth of percentage of Black people in the occupation.
For the rest of the minority groups, the fraction is partially dependent on the ground truth.
}
\label{fig:google_results}
\end{figure*}

\subsubsection{Gender labels}
 Overall, approximately 61\% of these images have a primary person whose gender is labelled as either Male or Female.
35\% of the images are labelled \textit{Male}, 26\% are labelled \textit{Female} and the rest are labelled ``Not applicable'' and ``Cannot determine''.
The variation of fraction of images of women in the results is presented in Figure~\ref{fig:google_results}a. 
The figure shows that Google images do follow the gender stereotype associated with occupations.
This was one of the main inferences of the case study by Kay et al. \cite{kay2015unequal}
 for Google 2013 search results.
While the overall fraction of women in the top 100 results seems to have increased from 2013 to 2019 (37\% in 2013 to 45\% in 2019), the fraction of gender anti-stereotypical images is still quite low (21\% in 2013 and 30\% in 2019).

\subsubsection{Skin-tone labels}
 For skin-tone, the options provided for labelling were the categories of Fitzpatrick skin-tone scale (Type 1-6). 
While there are more options in this case, choosing between consecutive options is relatively difficult.
Around 15\% of the images are assigned a Type-1 skin-tone label, 14\% Type-2, 5\% Type-3, 2\% Type-4, 2\% Type-5, 2\% Type-6; the rest are either  ``Not applicable'', ``Cannot determine'' or have conflicting skin-tone label responses.

However, our primary skin-tone evaluation is with respect to the fraction of images of dark-skinned people. Hence we can aggregate the skin-tones into a binary feature: \textit{fair} skin-tone (Type 1, Type 2, Type 3) and \textit{dark} skin-tone (Type 4, Type 5, Type 6).
After this aggregation, 52\% of the images are have fair skin-tone label and 10\% of the images have dark skin-tone label.
For the rest of the paper, we will treat the skin-tone as a binary feature, unless explicitly mentioned.

\begin{table}[t]
\centering
\small
\begin{tabular}{ |c|c |c |c |c| }
\hline 
\multicolumn{2} {|c|} { }  &  \multicolumn{2} {|c|} {\bfseries Diversity metrics} & \textbf{\bfseries Accuracy metric} \\ 
\cline{3-5}
  \multicolumn{2} {|c|} {Algorithm} & \makecell{\% gender \\anti-stereotypical} & \makecell{\% dark skinned} & avg. accuracy\\
\hline
\hline
\multirow{ 2}{*}{This paper} & QS-balanced & 0.45 (0.17) & 0.17 (0.05) & 0.38 (0.06) \\
& MMR-balanced & 0.45 (0.20) & 0.15 (0.06) & 0.39 (0.06) \\
\cdashline{1-5}
\multirow{ 7}{*}{Baselines} & QS & 0.35 (0.20) & 0.13 (0.06) & 0.47 (0.11) \\
& DS & 0.48 (0.20) & 0.15 (0.00) & 0.30 (0.06) \\
\cdashline{2-5}
 & Google& 0.30 (0.22) & 0.16 (0.09) & 0.48 (0.07) \\
& MMR & 0.35 (0.21) & 0.09 (0.05) & 0.48 (0.11) \\
& DET & 0.39 (0.15) & 0.15 (0.05) & 0.43 (0.08) \\
& AUTOLABEL & 0.36 (0.17) & 0.14 (0.05) & 0.47 (0.11) \\
& AUTOLABEL-RWD & 0.35 (0.21) & 0.13 (0.06) & 0.47 (0.11) \\\hline
\end{tabular}
\caption{Occupations dataset - {Comparison of top 50 images from \textbf{QS-balanced} and \textbf{MMR-balanced} algorithm with top 50 images from other baselines. }
The number represents the average, with standard deviation in brackets.
The accuracy is quantified using a measure of similarity to the query.
 \textbf{QS-balanced} returns an output set that has a larger fraction of images that do not correspond to the gender stereotype of the occupation. However, it suffers a loss in accuracy for this diversification. 
Note that accuracy is this case is measured using query similarity.
Other non-redundancy based algorithms also perform better than Google results in terms of gender diversity in the results, but not better than \textbf{QS-balanced} or \textbf{MMR-balanced}, showing that using the diversity control set targets the desired attributes better.}
\label{tbl:main_comparison}
\end{table}

\subsubsection{Intersection of gender and skin-tone}
 57\% of the images have both a gender and skin-tone (binary) label.
Amongst these, 27\% of the images are of fair-skinned men, 21\% are of fair-skinned women, 6\% are of dark-skinned men and 3\% are of dark-skinned women.
Once again, the fraction of images of dark-skinned men and women is relatively much smaller than the fraction of fair-skinned men and women, as seen from Figure~\ref{fig:google_results}b.
Furthermore, if we associate each occupation with its gender stereotype (for example, ``Male'' if the fraction of men in the occupation is larger than the fraction and women, and ``Female'' otherwise), then 35 out of 96 occupations do not have any images of dark-skinned gender anti-stereotypical people in the top 100 results.

Figure~\ref{fig:google_results}b also provides us with an insight into the variation of fraction of images of different groups (formed by intersection of gender and skin-tone) with respect to ground truth of fraction of Black people in occupations.
For almost all occupations, a large portion of the top 100 images are of gender stereotypical fair-skinned people, further showing that current Google results for occupations do correspond to the stereotypes.
Interestingly, the fraction of images of gender stereotypical fair-skinned people does not seem to be dependent on the ground truth.
While this partition takes up a significant portion of top 100 images, the fraction of images from other three minority partitions seem to be partially dependent on the ground truth.

This lack of gender diversity in Google results from 2013 has also been explored in detail in the paper by Kay et al. \cite{kay2015unequal}; our updated dataset shows that the current Google results still suffer from some of the gender diversity problems explored in that paper.
Furthermore, our analysis also shows that the Google image results are also lacking in terms of skin-tone diversity and intersectional diversity.

We will test the performance of \textbf{QS-balanced} and \textbf{MMR-balanced}
 algorithms on this Occupations dataset and compare the results, in terms of diversity and accuracy, to top Google results.

\subsection{CelebA dataset}
Another dataset we will use for evaluation is CelebA.
CelebA dataset \cite{liu2015faceattributes} is a dataset with 202599 images of celebrities, along with a number of facial attributes, such as whether the person in the image has eyeglasses or not, whether the person is smiling or not, etc.
We will use 37 of these attributes in our evaluation. One of the attributes corresponds to whether the person in the image is ``Male'' or not and we will use this attribute for diversity evaluation.

We divide the dataset into two parts: train and test set. 
The train set (containing 90\% of the images) is used to train a classification model over these attributes, which is then used to compute the query similarity score
The primary dataset for summarization is the test partition of the above CelebA dataset; it contains 19962 images.
The 37 facial attributes will serve as the queries to the summarization algorithm and the trained classification model will be used as the blackbox query algorithm $A(q, I)$.

{
Some of the attributes in this dataset are gender-neutral, while other seem to be gender-specific.
We consider an attribute to be gender-neutral if it is commonly associated with all genders and if the dataset has sufficient number of images from both men and women labelled with that attribute.
For example, we consider the attribute ``smiling'' to be gender-neutral since it is associated with both men and women, and amongst images labelled as smiling in the dataset, 34\% of images that are labelled as Male and 66\% are labelled as Female.
\footnote{Prior studies show that there is some correlation between gender and smiling for photographs taken during public occasions \cite{desantis2000women, dantcheva2016gender}.
However, summarization results should not reflect to bias of the source, i.e., when querying for a facial attribute like ``smiling'', which is associated with all genders, the results should be gender-diverse to present an unbiased picture.}
Similarly, the attribute ``eyeglasses'' can be considered gender-neutral since it is also commonly associated with both men and women, and the dataset has sufficient number of images of both men and women with eyeglasses.
On the other hand, an attribute like ``mustache'' is usually associated with men and all images labelled with this attribute in the dataset are of men; hence we will consider it to be gender-specific.
The fraction of images of women for other facial attributes are given in Section~\ref{sec:celeba_attr} in the Supplementary File.
Our primary goal for this dataset will be to ensure diversity with respect to such gender-neutral queries, but we will present our results for all the queries.
}

\begin{table}[t]
\centering
\small
\begin{tabular}{ |c|c |c |c |c| }
\hline 
\multicolumn{2} {|c|} { }  &  \multicolumn{1} {|c|} {\bfseries Diversity metric} & \textbf{\bfseries Accuracy metric} \\ 
\cline{3-4}
\multicolumn{2} {|c|} { Algorithm}  & \makecell{\% gender anti-stereotypical}  & avg. accuracy\\
\hline
\hline
\multirow{ 2}{*}{This paper} & QS-balanced &  0.23 (0.21) & 0.88 (0.16)\\
& MMR-balanced  &  0.17 (0.22) & 0.87 (0.16) \\
\cdashline{1-5}
\multirow{ 6}{*}{Baselines} & QS  &  0.08 (0.21) & 0.93 (0.16)\\
& DS  &   0.49 (0.12) & 0.22 (0.21) \\
\cdashline{2-4}
& MMR   &  0.14 (0.21) & 0.92 (0.16)\\
& DET   &  0.13 (0.18) & 0.90 (0.17)\\
& AUTOLABEL  & 0.5 (0)  & 0.80 (0.23)\\
& AUTOLABEL-RWD   &  0.07 (0.24) & 0.93 (0.17)\\
\hline
\end{tabular}
\caption{CelebA dataset - {Comparison of  top 50 images from all algorithms on the metrics of fraction of gender anti-stereotypical images, and accuracy.}
The accuracy is quantified as the fraction of images with corresponding query attribute.
The output returned by \textbf{QS-balanced} has a larger fraction of gender anti-stereotypical images than most of the other baselines.
Only \textbf{AUTOLABEL} returns a perfectly balanced set; however at a larger loss of accuracy.}
\label{tbl:celeba}
\end{table}

\section{Setup and observations} \label{sec:experiments}

We empirically evaluate the performance of \textbf{QS-balanced} and \textbf{MMR-balanced} on Occupations and CelebA dataset. 
The complete implementation details are provided in Section~\ref{sec:impl_details} in the Supplementary File, including the blackbox query algorithm and the similarity function used for each of the datasets; we provide certain important details of the implementation here.
In case of Occupations dataset, the query similarity is measured by quantifying similarity to a set of images corresponding to the query, while in case CelebA dataset, the query similarity is measured using output of a classifier pre-trained on the training partition of the dataset.

Since the choice of the diversity control set is dataset and domain-dependent, we discuss the content and construction of diversity control sets used for our simulations.
A detailed discussion on the composition, social and policy aspects of the diversity control sets is presented in Section~\ref{sec:div_control_discussion}..

\noindent
\subsection{Diversity control sets} \label{sec:div_control_set}
{
Our criteria to choose a diversity control set can be captured by the following points: (a) the diversity control set should consist of a small number of images that belong to the same domain as the dataset, and (b) the images should primarily differ with respect to the socially salient attribute and stay similar with respect to other attributes, such as background, face positioning, etc.
}

{
For the Occupations dataset, we evaluate our approach on four different small diversity control sets. Two sets (with 12 images each) are hand selected using images from Google results, and are intended to be diverse with respect to presented gender and skin color. 
The reason for using Google search to construct these sets was simply to ensure that the set is comprised of images from the same domain as the dataset itself.
These images are also not part of the Occupations dataset.
The other two sets (with 24 images each) are generated by randomly sub-sampling from the Pilot Parliaments Benchmark (PPB) dataset \cite{buolamwini2018gender}. 
We use PPB dataset to construct control sets because it contains portrait images of parliamentarians from different countries, and thus ensures that the images predominantly highlight the facial features of the person.
The images in the PPB dataset have gender and skin-tone labels, and we randomly select 24 images for our diversity control set, conditioned on the sampled set containing equal number of images of men and women and equal number of images of different skin-tones. These diversity control sets are presented in Section~\ref{sec:occ_div_control_set}.
}

{
For the CelebA dataset, once again we use four different diversity control sets for our evaluation, two of them have 8 images and the other two have 24 images; the exact images are provided in Section~\ref{sec:celeba_div_control_set}. The diversity control sets are constructed by randomly sampling equal number of images with and without the ``Male'' attribute from the train set.
Once again, we use the training part of the dataset to construct diversity control sets because, if possible, the images in the diversity control sets should be from the same domain as the dataset itself.
Since the domain, in this case, is images of celebrities, using images from training partition leads to better results (in terms of accuracy and diversity) than using images from Google search.
}

The results presented here compare the best performance using one of the diversity control sets and the comparison of different diversity control sets is presented in the Supplementary File.

\subsection{Baselines}
To better judge the results of our algorithms, we compare them to multiple other approaches as well as relevant baselines. 
We first consider two baselines that give the range of our options -- simply considering query accuracy (\textbf{QS}), or simply considering the diversity of the set (\textbf{DS}).
We also compare our results to the existing top Google results in the dataset.
{
For other baselines, we consider natural and effective approaches that have been proposed in prior image summarization literature.
To score images on query relevance, all algorithms once again either measure similarity using query images, in case of Occupations dataset, or use output of the trained classifier, in case of CelebA dataset.
To ensure diversity in the summary, prior work can be divided into two categories: algorithms that aim to reduce redundancy in the summary and algorithms that use socially salient attribute labels inferred using pre-trained classification tools.
We compare against both kinds of algorithms, and also discuss the potential drawbacks of these approaches below.
\subsubsection{Algorithms that ensure non-redundancy} Reducing redundancy is a common approach for achieving diversity in the output summary.
Essentially, algorithms that aim to maximize non-redundancy try to choose a summary which has images that are \textit{maximally-representative} of all the relevant images.
However, as shown by prior work \cite{celis2016fair} and our empirical results, this approach does not always effectively diversify across socially salient attributes, such as gender, and instead results in a summary that is diverse with respect to other attributes, such as background, body position, etc. 
We compare our algorithms against two approaches that fall under the category of reducing redundancy in the output summary.
\begin{itemize}
\item \textbf{DET}: Determinant-based diversification  \cite{kulesza2012determinantal, celis2018fair}. 
This approach first filters images according to their query relevance. Then it uses a geometric measure (determinant) on the features of a given subset of relevant images to quantify the diversity of the subset and aims to select the subset that maximizes this measure of diversity.
However, without any constraints on the subset, \textbf{DET} returns a summary that is diverse across all features, including irrelevant features such as background color, and hence can be unsuitable for the task of diversifying across the given socially salient attributes.
\item \textbf{MMR}: This algorithm is an iterative greedy algorithm that starts with an empty set and, in each iteration, adds an image that has \textit{maximum marginal relevance}, a score which combines both query relevance and extent of similarity to the images already chosen for the summary \cite{carbonell1998use}.
Similar to \textbf{DET}, we compare against this method to show that greedily choosing non-redundant images does not explicitly lead to diversity across socially salient attribute values.
\end{itemize}
\subsubsection{Algorithms that use label-inference tools} Many existing fair summarization algorithms assume the presence of socially salient attribute labels to generate fair summaries \cite{lin2011class, celis2018fair}, by using to labels to enforce \textit{fairness constraints} on the output summary.
In the absence of labels, one way to employ these algorithms is to use pre-trained classification tools to infer the socially salient attribute labels for all images in the dataset.
For example, one can use pre-trained gender classification tools to obtain gender labels for the images and then enforce constraints using these inferred labels.
However, this approach can be problematic if the classification model has been trained on biased data (as seen in \cite{buolamwini2018gender}) or has a relatively low accuracy for the given dataset.
In both cases, the use of a pre-trained gender classification model can further exacerbate the bias in the summary (as will be evident from empirical results on Occupations dataset).
For comparison of our approach against these kinds of methods, we use a pre-trained gender classification model \cite{levi2015emotion} and the following two approaches for generating summaries using query similarity scores and inferred labels.
\begin{itemize}
\item \textbf{AUTOLABEL: } Using pre-trained gender classification model \cite{levi2015emotion} \footnote{https://github.com/dpressel/rude-carnie}, this approach first divides the dataset into two partitions: images labelled ``male'' and images labelled ``female''.
Then it sorts images in each partition by query relevance score and selects equal number of top images labelled ``male'' and ``female'' for the summary.
\item \textbf{AUTOLABEL-RWD}: Once again using the same pre-trained gender classification model, along with a more effective scoring function suggested by \cite{lin2011class}; this approach rewards a subset for having images from multiple partitions instead of penalizing it for having images from the same partition.
\end{itemize}
Empirical comparison with these baselines will show that the bias or errors in pre-trained classification models can often exacerbate the bias of generated summaries or adversely affect their accuracy.
}

Additional mathematical details and descriptions of all the baselines are provided in Section~\ref{sec:baseline_details} of the Supplementary File.
Each algorithm, including the baselines, is used to create a summary of 50 images, corresponding to each query occupation.
{The comparison of our algorithms and baselines on smaller summary sizes is also presented in Section~\ref{sec:occ_diff_sizes} and \ref{sec:celeba_diff_sizes} in the Supplementary File.}
For Occupations dataset, we compare our algorithm and the baselines on metrics of gender diversity, skin color diversity and accuracy.
For CelebA dataset, we compare our algorithm and the baselines on metrics of gender diversity and accuracy.

\subsection{Observations - Gender diversity} \label{sec:results_gender}
\subsubsection{Occupations dataset}
As reported earlier, 52 out of 96 occupations are have a larger fraction of men employed and the rest have a larger fraction of women employed (inferred using the BLS data \cite{bls}).
We first report the fraction of gender anti-stereotypical images in the output for each query occupation, i.e., if an occupation is male-dominated, we take into account the fraction of women and if an occupation is female-dominated, we take into account the fraction of men in the output set.
The results are presented in Table~\ref{tbl:main_comparison}.
Algorithm \textbf{QS-balanced}, using PPB Control Set-1 returns a set for which the average fraction of gender-anti-stereotypical images is 0.45 with a standard deviation of 0.17.
In comparison, for Google Image search, the average fraction of gender-anti-stereotypical images in top results is 0.30 with a standard deviation of 0.22.
The table shows that \textbf{QS-balanced} algorithm returns a larger fraction of images that do not correspond to the gender-stereotype associated with the occupation.

In terms of raw gender numbers, the average fraction of women in top results of \textbf{QS-balanced}, for any occupation is 0.35 with a standard deviation of 0.10.
The results for performance of \textbf{QS-balanced} using other control sets is presented in the Section~\ref{sec:diff_control_sets_occ} of the Supplementary File.
Using Diversity Control Set-1 leads to a slightly larger average fraction of women; however using PPB Control Set-1 leads to better performance with respect to both gender and skin-tone, which is why we present our main results using this control set.

The gender diversity of the results of \textbf{MMR-balanced} is similar to those of  \textbf{QS-balanced}, and much better than Google results and baselines. 
The average fraction of gender anti-stereotypical images in the \textbf{MMR-balanced} is 0.45, with a standard deviation of 0.20, which is slightly worse than \textbf{QS-balanced} results. The average fraction of women in top results of any occupation for  \textbf{MMR-balanced} is 0.40 with a standard deviation of 0.17.
The results empirically show that the use of diversity control set appropriately, either in \textbf{QS-balanced} or \textbf{MMR-balanced}, leads to better diversification across gender.

\begin{figure*}[t]
\centering     
\subfigure[Gender diversity - Comparison with baselines]{
\includegraphics[width=0.48\linewidth]{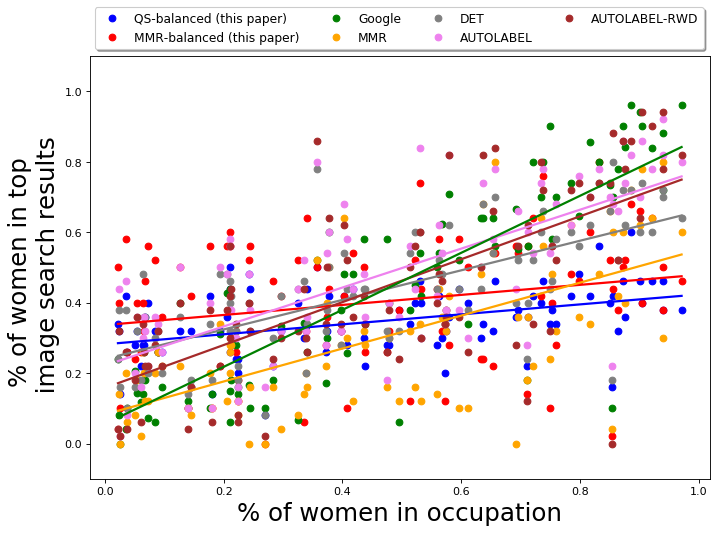}}
\subfigure[Skin-tone diversity - Comparison with baselines]{
\includegraphics[width=0.48\linewidth]{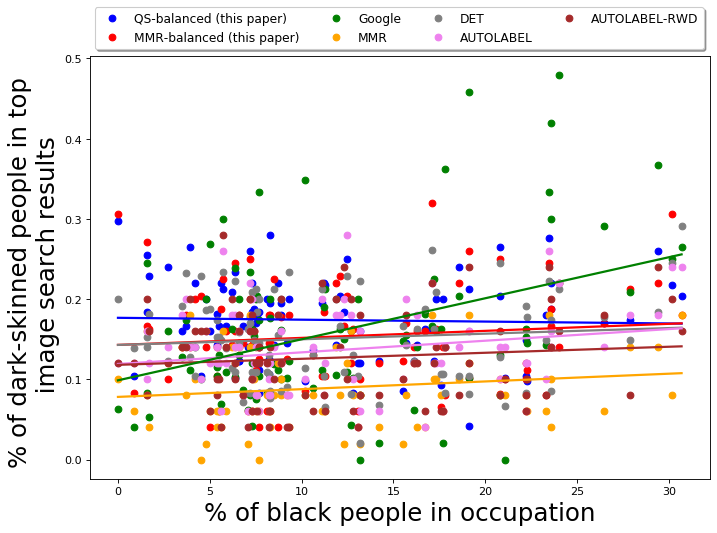}}
\caption{Occupations dataset: (a) Percentage of women in top 50 results vs ground truth of percentage of women in occupations. The images are generated using \textbf{QS-balanced}, \textbf{MMR-balanced} and other baselines for the Occupations dataset. 
The figure shows that the image results from \textbf{QS-balanced} and \textbf{MMR-balanced} are more gender-balanced (see also Table~\ref{tbl:main_comparison}), than image results from other algorithms.
While the fraction of images of women from \textbf{QS-balanced} is slightly lower than \textbf{MMR-balanced}, the fraction of gender-anti-stereotypical images for both algorithms is close (see Table~\ref{tbl:main_comparison}). 
(b) Percentage of dark-skinned people in top 50 results vs ground truth of percentage of Black people in occupations. 
The image results from \textbf{QS-balanced} are relatively more balanced with respect to skin-tone; however, the fraction of images of dark-skinned people is low for all algorithms.
}
\label{fig:main_comparison_2}
\end{figure*}

The variation of the percentage of women in the output of different algorithms is presented in Figure~\ref{fig:main_comparison_2}(a). 
The $x$-axis in Fig~\ref{fig:main_comparison_2}(a) is the actual percentage (ground truth) of women in occupations, obtained using data from BLS \cite{bls}.
The figure primarily shows the results from \textbf{MMR-balanced} and \textbf{QS-balanced} are relatively more gender-balanced,
On the other hand, \textbf{MMR} and \textbf{DET} have a relatively smaller fraction of gender anti-stereotypical images in their output.
This shows that algorithms that aim to diversify across feature space (like \textbf{MMR} and \textbf{DET}) cannot always achieve desired diversity with respect to the socially salient attributes, such as gender.
The fraction of gender anti-stereotypical is however better than Google results, showing that it does diversify across gender to an extent.

The performance of gender anti-stereotypical images in the output of \textbf{AUTOLABEL} and \textbf{AUTOLABEL-RWD} is relatively low as well (around 0.35); 
this is likely due to the low accuracy of the auto-gender classification tool used (error rate ~ 30\%).
The performance of these algorithms show that one cannot rely on automatic classification tools, for gender or other socially salient attributes, to ensure constraint-based diversification.
Hence, an intervention, in the form of a diversity control set, can help target the necessary attributes appropriately.

\subsubsection{CelebA dataset}
Table~\ref{tbl:celeba} shows that the output images of \textbf{QS-balanced} algorithm is more contains a larger fraction of gender anti-stereotypical images (0.23) than \textbf{MMR-balanced, MMR, DET, AUTOLABEL-RWD}.
The average loss in accuracy is also small (0.05) for \textbf{QS-balanced}.

On the other hand, the output set from \textbf{AUTOLABEL} algorithm is always perfectly balanced.
This is because the auto-gender classification tool used for CelebA dataset has a much better accuracy (95\%), and hence we are always able to choose a perfectly gender-balanced set.
However, the accuracy of this algorithm is relatively much worse than other algorithms; showing that enforcing hard fairness constraints does not always lead to the best results.

Even for image sets from \textbf{QS-balanced} and \textbf{MMR-balanced}, the overall fraction of gender anti-stereotypical images is not close to 50\%, as desired.
This is primarily because many queries correspond to a gender-stereotype; for example, most of the images satisfying the attribute ``wearing necklace'' correspond to female celebrities and hence the algorithm cannot diversify with respect to this feature, due to lack of images of men satisfying this attribute.
Similarly, most of the images satisfying the attribute ``bald'' correspond to male celebrities, and hence the images for this query mostly contain men.

On the other hand, our framework does lead to more gender-balanced results for queries that do not have an associated gender-stereotype.
For example, for the query ``smiling'', the top 50 images with the best query scores contains only images of women, whereas the results from \textbf{QS-balanced} contain around 36\% men and 64\% women images.
Similarly, for the query ``receding hairline'', the top 50 images with the best query scores contains 12\% women, whereas \textbf{QS-balanced} returns an image set with 38\% women.
Hence, for queries which are gender-neutral, using our framework leads to results that are relatively more gender-balanced.

\subsection{Observations - Skin-tone diversity}

\subsubsection{Occupations dataset}
Unlike gender, for skin-tone, dark-skinned people are the minority group for all occupations considered in this dataset.
Hence, in this case the fraction of anti-stereotypical images just corresponds to fraction of images of dark-skinned people.

Using Algorithm \textbf{QS-balanced}, with PPB Control Set-1, the average fraction of people with dark skin-tone in top results of any occupation is 0.17 with a standard deviation of 0.05;
for Google Image search, the average fraction of women in top 50 results for any occupation is 0.16 with a standard deviation of 0.09.
The high standard deviation shows that Google results are relatively more imbalanced with respect to gender, i.e., for many occupations, the fraction of images of dark-skinned people is much smaller or larger than the average.
The skin-tone diversity of the results of \textbf{MMR-balanced} is also relatively better than baselines; the average fraction of women in top results of any occupation is 0.15 with a standard deviation of 0.06.

We also compare the skin-tone diversity of results of  \textbf{QS-balanced}  with other baseline algorithms; the results are presented in Table~\ref{tbl:main_comparison} and Figure~\ref{fig:main_comparison_2}(b). 
The $x$-axis in Fig~\ref{fig:main_comparison_2}(b) is the actual percentage (ground truth) of Black people in occupations, once again obtained using data from Bureau of Labor and Statistics \cite{bls}.

Once again \textbf{MMR} is unable to diversify across the desired attributes.
For the results obtained using \textbf{MMR}, the average fraction of people with dark skin-tone in top results is 0.09, with a standard deviation of 0.05.
The skin-tone diversity of results of \textbf{DET} is relatively better, the average fraction of people with dark skin-tone in top results is 0.15, with a standard deviation of 0.05.

Note that for all algorithms, the top results still have a very small fraction of people with dark skin-tone (despite using a diversity control set that is balanced with respect to skin-tone).
This is primarily because, for most occupations, there are very few images of people with dark skin-tone in the dataset.
We expect that summarization over a more robust dataset (such as one accessible to Google for search results) can lead to better results.

\subsection{Intersectional diversity}
In the presence of multiple socially salient attributes, intersectional diversity would imply that the results are diverse with the respect to the combination of the socially salient attributes.
\subsubsection{Occupations dataset }
We evaluate the performance of \textbf{QS-balanced algorithm} on the basis of intersectional diversity with respect to gender and skin-tone attributes.
In other words, we check how the output set is distributed across the following four partitions: gender stereotypical fair skin-tone images, gender anti-stereotypical fair skin-tone images, gender stereotypical dark skin-tone images and gender anti-stereotypical dark skin-tone images.
The results are presented in Table~\ref{tbl:int_div}. The diversity control set used here is PPB Control Set-1.

As discussed earlier, Google images tend to favor the gender and skin-tone associated with the stereotype of the occupation; the table shows that the fraction for gender-stereotypical fair skin-tone images is much larger than the fraction for other partitions.
In comparison, the results from \textbf{QS-balanced} are relatively more balanced; the difference between the fraction of gender-stereotypical and gender anti-stereotypical images is smaller, for both fair skin-tone and dark-skin tone.
Furthermore, the fraction of gender anti-stereotypical dark skin-tone images in the output of \textbf{QS-balanced} is also larger than the corresponding fraction in Google images.
The comparison with other baselines is also presented in Table~\ref{tbl:int_div_all} in the Supplementary File.

Overall, the fraction of  gender anti-stereotypical dark skinned images is still low in the output of \textbf{QS-balanced}. Once again, the primary reason for this is the lack of robustness of the dataset itself. As noted earlier, for 35 occupations, the dataset does not contain any  gender anti-stereotypical dark skinned images; to choose such images for these queries, the algorithm has to look for similarity with images from other occupations, which leads to a small fraction of gender anti-stereotypical dark skinned images and also affects accuracy.

\subsection{Observations - Accuracy}
\subsubsection{Occupations dataset}

For the Occupations dataset, we compute accuracy by measuring similarity to the query in the following manner:
{for every query occupation $q$, we have a small set of images $T_q$ for reference; for example, for query ``doctor'', 10 images of doctors are provided. \footnote{{These images are hand-verified and are not present in the primary evaluation dataset $S$}}. 
Then using $\sim$ function, for the reference set $T_q$ and for each image $I$ in summary, we can calculate the score $\avgSim_{T_q}(I ) := \avg_{I_q \in T_q} sim(I , I_q )$. The score $\avgSim_{T_q} (I)$ gives us a quantification of how similar the image $I$ is to all other images in set $T_q$, and correspondingly how similar it is to query $q$.}\footnote{{This is similar to the ROUGE score \cite{lin2003automatic} employed to measure the utility of text summaries against reference summaries and has been shown to correlate well with human judgment.}}
The query similarity of different algorithms and baselines is presented in Table~\ref{tbl:main_comparison}.
\footnote{For Occupations dataset, we can also alternately define accuracy as the fraction of images in the summary that belong to the query occupation.
However, this measure is problematic since many occupations have similar looking images, for example, ``doctor'' and ``chemist'', or ``insurance sales agent'' and ``financial advisor''.
Hence, similarity with reference images is a better measure of accuracy in this case; nevertheless, we also present the accuracy with respect to query occupation in Section~\ref{sec:occ_acc_alt} of the Supplementary File.}

From the figure, we can see that the accuracy of the top images of \textbf{QS-balanced} (0.38) and \textbf{MMR-balanced} is relatively lower than the top images of Google image search (0.48).
The average accuracy of other baselines is slightly better than our primary algorithms (greater than 0.42).
Hence the loss in accuracy, due to the incorporation of diversity control matrix, is not very large.

Note that query similarity does not imply that most of the output images belong to the query occupation.
There will be images from other occupations that are matched to the query occupation, since multiple occupations can have similar images (for example, doctors and pharmacists, or CEOs and financial analysts).
The plot presented here simply checks whether the average query scores of the output images of  \textbf{QS-balanced}  and \textbf{MMR-balanced} are close to the Google search results and other baselines.
To further check the number of images in the output set that belong to the query occupation, we plot a bar graph of number of images belonging to the query occupation and the results are presented in in Figure~\ref{fig:acc_bar_graph} in the Supplementary File.

\subsubsection{CelebA dataset}
Table~\ref{tbl:celeba} also shows the accuracy comparison of our algorithm on CelebA dataset against baselines.
Here the accuracy is measured as the fraction of images that satisfied the query facial attribute.
As expected, the accuracy of the results when using \textbf{QS-balanced} (88\%) is worse than the accuracy when \textbf{QS} (93\%), but better than the average accuracy of\textbf{DS} (22\%), \textbf{MMR-balanced} (87\%) and \textbf{AUTOLABEL} (80\%).
The reason for relatively lower accuracy of \textbf{MMR-balanced} is primarily because it aims to reduce non-redundancy in the summary as well.

For some queries, such as ``smiling'' or ``eyeglasses'', the loss in accuracy is small (2\%), while for other queries, such as ``straight hair'', even though the accuracy is small (72\%), the images do visually correspond to the query.
For these kinds of queries, the performance of our algorithm (in terms of accuracy and diversity) seems to be as desired.
For some other queries, such ``mustache'' or ``wearing lipstick'', the use of diversity control scores with $\alpha-0.5$ does not seem to have an impact of gender diversity (0\% gender anti-stereotypical images for both).
This is primarily because these queries are associated with a gender-stereotype, in which case forced diversification will affect accuracy.

\subsection{Observations - Other diversity metrics}
{
We also evaluate the performance of \textbf{QS-balanced}, \textbf{MMR-balanced} and baselines with respect to other standard diversity metrics from literature, such as non-redundancy scores (measured using log-determinant of the kernel matrix). 
The details and results of this comparison are presented in Section~\ref{sec:non_red_appendix} in the Supplementary File.
To state the observations briefly, the non-redundancy scores of the output generated by \textbf{DET} are observed to be better than non-redundancy scores of other algorithms. This is expected, since \textbf{DET} optimizes the determinant-metric being measured.
However, as noted before, maximizing non-redundancy does not necessarily ensure diversity with respect to gender and skin-tone.
Amongst the algorithms from this paper, \textbf{MMR-balanced} has relatively better non-redundancy scores than \textbf{QS-balanced}.
This is primarily because \textbf{MMR-balanced} and has a non-redundancy component already built into it (at the cost of efficiency); \textbf{QS-balanced}, on the other hand, is faster since it only aims to ensure diversity with respect to attributes represented in the control set.
}

\section{Discussion, limitations and future work} \label{sec:limitations}
The algorithms presented here are prototypes that aim to improve diversity in image summarization. 
A crucial feature of our framework is that it is built to extend existing image summarization algorithms (represented using the blackbox $A(\cdot, \cdot)$).
This is because summarization algorithms can be designed in a manner very specific to the domain; for example, Google Image search uses metadata of the images (such as parent website, website metadata, etc) to return images that correspond to the query.
Designing a new fair summarization from scratch is unreasonable, and a post-processing approach to ensuring fairness is more likely to be adopted.
However, there are certain limitations to this approach which we examine in connection to potential future work in this section.

\subsection{Discussion on the observations}
The empirical results show that using the diversity control set has a positive impact on the gender and skin-tone diversity of the summary, either in the form of \textbf{QS-balanced} or \textbf{MMR-balanced} algorithm.
The average fraction of gender anti-stereotypical images in the output of both algorithms is close to 0.45, for the Occupations dataset.
In comparison, the average fraction of gender anti-stereotypical images in Google images is around 0.30.
Even the algorithms that aim to just reduce non-redundancy, are unable to diversify across gender and skin-tone to the extent that  \textbf{QS-balanced} or \textbf{MMR-balanced} does.

However, the results for skin-tone and intersectional diversity of the results of \textbf{QS-balanced} and \textbf{MMR-balanced} on Occupations dataset is still lower than the desired level of diversity (close to the fraction in the control set).
Even though this is because of the lack of images of people with darker skin-tone in the Occupations dataset, it will be to important to empirically evaluate the performance of the framework on more robust datasets. 

In case of CelebA dataset, while the overall average fraction of gender anti-stereotypical images is not very high (0.23), we do observe that for certain queries, the fraction of gender anti-stereotypical images is higher than those obtained using just query scores (for example, ``smiling'').
These queries mostly correspond to gender-neutral facial attributes, for which there are sufficient images in the dataset.

\subsection{Comparison with baselines}
From the performance of \textbf{DET} and \textbf{MMR}, we see that diversifying across feature space does not necessarily diversify across the socially salient attributes; an observation also made in \cite{celis2016fair}.
Furthermore, imposing hard fairness constraints (such as using \textbf{AUTOLABEL} when the pre-trained gender classifier has high accuracy) is not ideal since this can lead to undesirably high loss of accuracy.
Hence diversity control sets can serve as a medium of \textit{soft fairness constraints}.
 
\subsection{Diversity control sets} \label{sec:div_control_discussion}
While diversity control sets, when appropriately chosen, do seem to improve the diversity of the output, the choice of the composition of diversity control set is context-dependent.
It is obvious that the diversity control set images should be chosen keeping in mind the domain of the images of the dataset, to ensure that image similarity comparison is not redundant.
For example, a visible age-gap between the people in the images in the dataset and the control-set can only lead to more inaccurate results.
Even the diversity control sets used for Occupations dataset do not perform well for CelebA dataset, showing that the choice of control set has to be dataset-specific.

But what should be the fraction of images of women or dark-skinned people in the diversity control set?
We observe that changing the composition of the diversity control set changes the composition of the output similarly.
We infer this by empirically evaluating the performance of \textbf{QS-balanced} algorithm for diversity control sets with different fractions of images of minorities and observe that as the fraction increases, the representation of images of these minorities in the output set also increases.
The diversity control sets are randomly chosen from the PPB dataset.
The results for this analysis are presented in Section~\ref{sec:diff_comp_occ} of the Supplementary File.
Hence, the composition of the diversity control set does seem to have an impact on the composition of the output summary.

The size of the control set is intentionally kept to be very small (recall that the time complexity depends linearly on the size of the control set). 
Indeed it is a key advantage of our approach that it performs well even with small control sets. 
Larger control sets could be used, but constructing them could be considerably more difficult, especially considering that determining the control set is context-specific and could/should require input from multiple parties. 
Empirically, we did not observe any statistically significant advantage in using a control sets of size 100-200.

There are many other context-specific and policy-related questions about the diversity control set that cannot be answered through the above empirical analysis.
Typically for an application, the range of composition of the control set should be decided after a thorough research of the user demographics and will also require input from all the affected parties/communities to ensure that there is an appropriate representation of all groups.  
Once the control set is created and deployed, ideally the company responsible for the application of the framework should also provide opportunities for public audit/examination of the criteria and diversity sets to ensure transparency in the diversification process.
The reason why transparency is required in the process of selection of a diversity control set is because, just like any other fairness metric, using misrepresentative or non-diverse control sets can lead to more harm than good. 
Similar to the process adopted in other settings such as voting \cite{valais}, it should be upto to the user-base to decide/judge the fairness of a diversity control set.

\subsection{Choice of tradeoff parameter $\alpha$}
The hyper-parameter $\alpha$ represents the fairness-accuracy tradeoff in this algorithm.
Once again, the choice is application oriented and depends on how much loss in accuracy is acceptable to achieve the required amount of fairness in the output.
We empirically evaluate the performance of \textbf{QS-balanced} and \textbf{MMR-balanced} for different $\alpha$ values, and the results are presented in Section~\ref{sec:diff_alpha_occ}  and \ref{sec:diff_alpha_celeba} of the Supplementary File.
As expected, as $\alpha$ increases from 0 to 1, the fraction of gender anti-stereotypical images (for both Occupations and CelebA dataset) increases. 
At the same time, the similarity to the query or accuracy decreases.
In our case, the figures show that a balanced choice of $\alpha = 0.5$ is reasonable.

The choice of hyper-parameters, such as diversity control set and $\alpha$ value are context-dependent and we expect the use of this algorithm to be preceded by a similar thorough evaluation and analysis using different control sets with different compositions, and different $\alpha$ values.

\subsection{Socially salient attributes}
It is important to state that the primary evaluation of our method was with respect to gender. 
This evaluation made use of labelled data where gender was primarily treated as binary, which is unnecessarily and problematically restrictive \cite{julia2016misgendering}, not an accurate representation of the gender diversity in humanity \cite{gherovici2011please}, and can be used in a discriminative manner \cite{bender2016peeing, herman2013gendered}.
It would be important to evaluate this work in light of other socially salient attributes, and broader label classes. 

The lack of analysis and evaluation with respect to non-binary gender attribute is a limitation of many existing gender-classification tools as well.
A study conducted by Scheuerman, Paul and Brubaker \cite{scheuerman2019computers} showed that existing commercial facial analysis tools do not perform well for transgender individuals and are unable to infer non-binary gender, primarily because of focus of training on recognizing gender-stereotypical facial features.
Such studies further highlight the importance of not relying on the pre-defined notion of gender, as considered by existing gender classification tools.

\subsection{Better implementation techniques}
Despite the diversity control sets being balanced across male/female presented genders, the results from  \textbf{QS-balanced} do not match these ratios exactly, and there is scope for improvement, perhaps with better diversity sets or similarity functions. 
Our current query matching algorithm for Occupations dataset  is based only on the similarity with the query control set images and can be improved given additional information about the image.
Once again, for a model similar to Google Image search, one would have access to the metadata of the image which will help better quantify query similarity or the similarity of two images.
Other transfer learning techniques, like retraining a small part or a single layer of the CNN, could also be employed for better feature extraction, although we did not see any improvement on an initial approach in this direction.

Just like other aspects of our algorithms, the implementation will also be context-specific.
For example, in case of CelebA dataset, we had a highly-accurate multi-class classifier to determine query similarity. 
Hence, in this case the accuracy of the output summaries was quite high (in the range of 85\% to 90\%).
On the other hand, for Occupations dataset, we had to use a generic similarity measure (average similarity with query images), which cannot be expected to have the best performance for every dataset.

\subsection{Evaluation in the absence of labels}
Another challenge of using this approach is that it may not always be easy to evaluate its success. 
Its main strength -- that it can diversify without needing class labels in the training data -- is also an important weakness because we may not always have labelled data with which to evaluate the results. 
One approach would be to predict labels using, e.g., gender classification tools \cite{levi2015emotion}.
However, we do not recommend using predicted labels in general as such classification tools can themselves introduce biases (as seen with the baseline \textbf{AUTOLABEL} for Occupations dataset) and are currently not designed with broader label classes or non-binary gender in mind, and hence do not address the core problem.
Perhaps a better approach would be to use human evaluators to rate or define the visible diversity of the images selected by the algorithm. 

{
The absence of labels also limits of our analysis to relatively-small datasets.
Real-world image datasets handled by applications like Google Search are considerably larger than the ones used in this paper and are often handled as data streams \cite{mirzasoleiman2018streaming,feldman2018less}.
However, without socially salient attribute labels, the diversity of summaries for large datasets cannot be evaluated.
At the same time, since the application of our framework is independent of the labels, the performance reported in this paper should extend to larger datasets as well, and as part of future work, exploring techniques to evaluate performance on large datasets will help  establish the scalability of our approach.
}

\subsection{Community-driven application of the framework} 
{
Our work can also been seen in the light of the push towards participatory technologies in machine learning.
Uninformed application of any technology that aims to ensure fairness can inadvertently cause  more harm than good \cite{liu2019delayed, zhang2019long, bennett2019point}.
Recent studies exploring the current and future applicability of works in fairness literature have correspondingly emphasized on the importance of participation of all stakeholders in the design-process of an application \cite{sassaman2020creating, chancellor2019relationships, muller2007participatory, disalvo2012communities}.
Such a design process is especially important for summarization models since the results of these models can shape the perceptions of the users.
Participatory design encourages the practitioners to engage with the users of the application to obtain valuable feedback on the possible disparate impacts of the application and ensures that there is a balanced power relation between the user and the engineer designing of an application \cite{sanders2002user, muller2007participatory, friedman2008value, le2009values}.

An important aspect of our framework is that it requires community participation to ensure its success.
As discussed in Section~\ref{sec:div_control_discussion}, the selection of a diversity control set should regularly take user feedback into account to guarantee that it is sufficiently representative of the user demographics.
Encouraging community participation also ensures that the decisions regarding key aspects of the summarization framework are not entirely made by engineers.
Crucially this shifts the power of the design process away from organizations and applications like Google Search and towards the users affected by the search results.

Furthermore, 
a crucial advantage of our framework is its post-processing nature; given any existing blackbox summarization or ranking algorithm, our framework adds a diversification component above the blackbox algorithm to ensure that the summary is fair; hence the implementation of the framework can be independent of the organization responsible for the blackbox algorithm.
This advantage can be exploited in settings where the blackbox algorithm cannot be modified. For example, our framework can possibly be implemented as a browser extension or a separate web-application created by a third-party that uses results from Google Image Search API and maintains a diversity control set.
However, the absence of participation of the organization that designed the blacbox summarization algorithm may also not be ideal.
The engineers who design the summarization algorithm would have considerably more knowledge of the domain of the datasets and can better decide the feasibility of any control set, as well as, its impact on the accuracy of the results.
As discussed earlier, an inappropriately chosen control set can lead to exacerbation of biases in the output generated by the framework, and to prevent this, one has to make sure that the control set images belong to the same domain as the dataset.
Given that the users only see a fraction of the dataset at any point of time, they cannot be expected to accurately judge the feasibility of any control set.
The ideal use of diversity control sets would, therefore, need involvement and discussion from all parties.
Importantly, our framework provides an opportunity for such a discussion and can help create a balanced power dynamic between the designers of search algorithms and the users of these algorithms, when deciding how well the results should represent the user demographics.

}

\newpage
\section{Conclusion}
The approaches presented in this paper (\textbf{QS-balanced} and \textbf{MMR-balanced})  aim to ensure fairness in summarization algorithms in the absence of (explicitly or inferred) labelled data in either training or deployment in order to decrease bias.
%
As a post-processing approach, it is also flexible in that it can be applied post-hoc to an existing system where the only additional input necessary is a small set of ``diverse'' images. 
We show its efficacy on two datasets: an image dataset of occupations where it can significantly improve the diversity of the images selected with little cost to accuracy as compared to images selected by Google, and an image dataset of celebrities where it can selects significantly more diverse images for gender-neutral facial attributes.
Due to the generality and simplicity of the approach, we expect our algorithm to perform well for a variety of domains, and it would be interesting to see to what extent it can be applied in areas beyond image summarization. 
\bibliographystyle{plain}
\bibliography{references}
\newpage

\appendix

\section{Details of baselines} \label{sec:baseline_details}
In this section, we provide the details of the baselines that we compare our algorithms against.
The first is determinant-based diversification  \cite{kulesza2012determinantal, celis2018fair}, \textbf{DET}. 
This approach effectively diversifies the selected images across their feature space.
Suppose that we need to return $M$ images corresponding to the query $q$.
Given the query similarity scores $A(q,I)$, we can sort the list in ascending order and extract the first $c \cdot M$ images from the list, where $c >1$ (we use $c \approx 3$ in our experiments), denoted by $\mathcal{W}_{c,q}$.
We can then employ the following standard diversification technique to find the most diverse images in the set $\mathcal{W}_{c,q}$.
For any $W \subseteq \mathcal{W}_{c,q}$, such that $|W| = M$, let $V_W$ denote the matrix with the feature vectors of images in $W$ as rows.
Then return the set $$\arg \max_{W \subseteq \mathcal{W}_{c,q}} \det(V_W V_W^\top).$$
If the number of subsets $W$ is large (can be exponential), we use greedy approximate algorithms for this task \cite{nemhauser1978analysis}.

Next we compare with respect to another algorithm that aims to reduce redundancy in the final set, \textbf{MMR}.
The algorithm is an iterative algorithm that starts with an empty set $R$ and adds one image to $R$ in each iteration.
The chosen image is the one that minimizes the score
\[ \alpha \cdot A(q, I) - (1 - \alpha) \cdot \min_{J \in R} \sim(I,J).\]
The first part of the above expression captures query relevance while the second part penalizes an image according to similarity to existing images in the summary $R$.
This algorithm (also referred to as \textit{maximum marginal relevance}) is a popular document summarization algorithm to reduce redundancy \cite{carbonell1998use}.
We will use $\alpha = 0.5$.

The baselines \textbf{DET} and \textbf{MMR} aim to show the importance of having a diversity control set.
In the absence of any attribute information with respect to which the results are expected to be diverse (for example, say gender), directly diversifying the output images will results in images that are diverse in unimportant features like background etc.
The diversity control set $T_F$ helps us identify the features for which diversity should be ensured.

For the third and fourth  baseline, we will use automatic gender classification tools.
Using existing pre-trained gender classification models, in particular \cite{levi2015emotion} \footnote{https://github.com/dpressel/rude-carnie}, we derive the gender labels for the images in the small dataset.

The third baseline, \textbf{AUTOLABEL}, is the following: we select $M/2$ images labelled \textit{male} (by the classification tool) with the best query relevance score $A(q,I)$ and $M/2$ images labelled \textit{female} with the best query relevance score $A(q,I)$.
For evaluation, however, we use the true gender labels of the images.
The purpose of this baseline is to show that using existing \textit{imperfect} auto-labelling tools to set constraints for diversification can lead to magnifying the biases already present in the pre-trained classification model used.

For the fourth baseline \textbf{AUTOLABEL-RWD}, we use the monotone submodular function proposed by \cite{lin2011class}.
They suggest that instead of penalizing a subset for having redundant images, one should reward a subset for being diverse.
The scoring function to measure the quality of a set $R$ is then the following (adapted for our domain):
\[\textrm{rwd}(R) := \sum_{I \in R} A(q, I) + \sum_{i=1}^K \sqrt{\sum_{I \in R \cap P_i } A(q,I) },\]
where $P_1, \dots, P_K$ are the partitions of the domain based on the socially salient attribute. 
For the case of gender, we will have two partitions.
The second part of the expression ensures that adding images from different partitions has a higher diversity score than adding images from same partition.
Once again, we will create the partitions according to the gender labels obtained using the classification tool.
We will use a greedy algorithm to obtain an approximately optimal subset for this case, since finding the optimal solution directly has a large time complexity.
The greedy algorithm will simply add the image $\arg \min_{I \in S \setminus R} \textrm{rwd}(R \cup \br{I})$ at every step, where $R$ is the subset chosen so far.

\section{Implementation details} \label{sec:impl_details}
In this section, we provide the complete implementation details, starting with the query matching algorithm $A(\cdot, \cdot)$ and the similarity function $\sim(\cdot, \cdot)$ used in our empirical analysis.

\noindent
\subsection{Image similarity}
To obtain the similarity score $\sim(I_1, I_2)$ for two given images, we can utilize a pre-trained convolutional neural network. 
We use the VGG-16 network \cite{simonyan2014very}, a 16-layer CNN, pre-trained on Imagenet \cite{imagenet} dataset, for generating the feature vectors\footnote{other networks such as \cite{schroff2015facenet} could similarly be used instead.}. 
We take the weights of the edges from the last fully-connected layer as the feature vector for the image.
The process can be summarized in the following steps \footnote{Similar to one-shot learning using Siamese Networks \cite{koch2015siamese}.} 
\footnote{Cosine distance has been used in document summarization literature to calculate similarity \cite{lin2011class} as well.
The cosine distance metric also outperformed other norm-based metrics (see Fig~\ref{fig:accuracy_1_norm} in the Appendix).}: 
(1) feed the image $I_1, I_2$ into the VGG-16 network and obtain the feature vectors $v_{I_1}, v_{I_2}$ of dimension 4096,
(2) perform Principal Component Analysis to reduce the feature vector size,
(3) return the cosine distance as similarity score, i.e., 
$$\sim(I_1, I_2) = 1 - \frac{ v_{I_1} \cdot v_{I_2}}{\norm{v_{I_1}}_2 \norm{v_{I_2}}_2}.$$

This method of using pre-trained models for other tasks is also called ``transfer learning''.
This technique has been succesfully employed in many other image-related tasks \cite{oquab2014learning}.

\noindent
\subsection{Query matching}
 \textbf{QS-balanced} (Algorithm~\ref{algo:main}) and \textbf{MMR-balanced} use a black-box querying algorithm $A$ to rank images according to similarity to a query.
For evaluation purposes, we describe an algorithm for query matching algorithm in case of Occupations and CelebA datasets.

\noindent
\textbf{Query matching algorithm $A$ for Occupations dataset:}
Suppose that for every $q$, we are provided a small set of images $T_q$; for example, for query ``doctor'', 10 images of doctors (that can be hand-verified).
Then using $\sim$ function, for the query set $T_q$ and for each image $I \in S$, we can calculate the score
$\textrm{avgSim}_{T_q}(I) := \avg_{I_q \in T_q} \sim(I, I_q).$
The score $\textrm{avgSim}_{T_q}(I)$ gives us a quantification of how similar the image $I$ is to all other images in set $T_q$, and correspondingly how similar it is to query $q$. 
Before using this score further, we can normalize it by subtracting the mean and dividing by standard deviation.
Therefore given a set $T_q$, for each $I \in S$, the query similarity score can be defined as
$$A(q ,I) := \widehat{\avgSim}_{T_q}(I) = \frac{\avgSim_{T_q}(I) - \textrm{mean}(\avgSim_{T_q})}{\textrm{std}(\avgSim_{T_q})}.$$
We will use this score to compute $\DS_{I_F}^q$.

For each query occupation $q$, we use the top 10 images from Google results of that occupation in the dataset as the similarity control set $T_q$.
Note that we use this query relevance algorithm for other baselines which employ $A(q, \cdot)$ score as well.

When we have to report accuracy for results over Occupations dataset, we will use the measure of query similarity.
While the above score $\widehat{\avgSim}_{T_q}$ is a measure of query similarity, is represents high similarity if the value is lower.
To avoid confusion, and maintain the convention that high value is high accuracy, when measuring accuracy we will use $\sim(I_1, I_2) = \frac{ v_{I_1} \cdot v_{I_2}}{\norm{v_{I_1}}_2 \norm{v_{I_2}}_2}$ and then similarity calculate average similarity with respect to all query images.

\noindent
\textbf{Query matching algorithm $A$ for CelebA dataset:}
For the CelebA dataset, recall that we divide the dataset into train and test partitions.
The train partition is used to train a multi-class classification model, with the facial attributes as the labels.

The classification model, given an input image, returns a vector of length 37, where each entry ($\in [0,1]$) represents the probability that the input image satisfies the corresponding attribute; let $f : S \rightarrow [0,1]^{37}$ denote the classifier.
We use the MobileNetV2 architecture and a transfer learning approach suggested by Anzalone et al. \cite{anzalone2019transfer} for the classifier, which achieves a training accuracy of around 90\%.

Since we follow the convention that the smaller the score the better the image corresponds to the query, we will use the negative of the classifier output as the query-similarity score, i.e., $A(q, I) = -f^{(q)}(I)$, where $f(I)$ denotes the output of classifier for image $I$ and $f^{(q)}$ denotes the entry corresponding to the attribute $q$.

For the image-similarity scoring function, we will use the pre-trained VGG-16 network to extract the features of the images and return the cosine distance between the features as the similarity score between the images.

\subsection{Diversity Control Matrix}
Finally, to efficiently implement  \textbf{QS-balanced}, we can construct a diversity control matrix of size $|S| \times |T_F|$ using the image-similarity scores between the images in $S$ and images in $T_F$. 
Before using this matrix to compute the $\DS_{I_F}^q$ scores, we will normalize each column of this matrix, i.e., we compute $\widehat{\avgSim}_{\br{I_F}}(I)$.
Therefore the final $\DS_{I_F}^q$ score is evaluated as 
\[\DS_{I_F}^q(I) =          \alpha \cdot  \widehat{\avgSim_{\br{I_F}}(I)} + (1 - \alpha) \cdot \widehat{\avgSim_{T_q}(I)}.\]
To implement this approach efficiently, we calculate the scores $\sim(I,I_F)$ as a pre-processing step and store them in the diversity control matrix.
Then given a query, we calculate the scores $\widehat{\avgSim_{T_q}(I)}$ and combine the diversity  control matrix and query similarity score list to get a matrix of size $|S| \times |T_F|$, where the element corresponding to $I \in S$ and $I_F \in T_F$ has the value $\DS_{I_F}^q(I)$.

For \textbf{MMR-balanced}, we use the greedy approach and add the diversity score of an image to its relevance score at every step.

\section{Model Properties} \label{sec:discussion}
As mentioned earlier in the Related Work section, query-based diverse summarization has been a major area of research in many sub-domains within information retrieval.
For diverse document and image summarization, multiple models have been considered and evaluated rigorously \cite{carbonell1998use, lin2011class, tschiatschek2014learning}.
Some of the models we consider as baselines are derived from models that are popular and commonly used in diverse document summarization literature (\textbf{MMR}, \textbf{DET} and \textbf{AUTOLABEL-RWD}).
One of the properties that a lot of diversity-ensuring summarization models share is the property of submodularity, defined formally below.
\begin{definition}[Submodular function] Given a set of elements $\Omega = \br{x_1, \dots, x_n}$ and a function $f : 2^\Omega \rightarrow \R$, the function $f$ is called submodular if it satisfies the property that for any $R_1 \subseteq R_2 \subseteq \Omega$ and any element $x \in \Omega$,
\[f(R_1 \cup \br{x}) - f(R_1) \geq f(R_2 \cup \br{x}) - f(R_2).\]
\end{definition}
\noindent
Submodular functions quantify the property of \textit{diminishing returns}, and in many settings, a simple greedy approach can return a good approximation of the optimal solution for maximizing a submodular function.
In case of maximizing monotone submodular functions subject to cardinality/matroid constraints, a greedy algorithm returns a 0.632-factor approximation to the optimal subset in the worst case \cite{nemhauser1978analysis, calinescu2007maximizing}, and in many cases, performs much better than the worst-case bound.

Submodular functions occur naturally when the task is to ensure that the output summary is representative of a particular subdomain of the population.
For example, in case of image summarization tasks that aim to reduce redundancy or ensure representativeness in the final set, Tschiatschek et al. \cite{tschiatschek2014learning} argued that many models in existing literature are cases of submodular maximization.
Even algorithms based on determinantal point processes, such as \cite{celis2018fair, kulesza2012determinantal} satisfy this property, since the determinant-based objective function is log-submodular.

In this section, we show that the scoring mechanisms considered in Section~\ref{sec:experiments} satisfy the diminishing returns property and are in-line with the submodularity property common to the diverse summarization literature.
The submodularity of \textbf{MMR}, \textbf{AUTOLABEL-RWD} \cite{lin2011class} and \textbf{DET} \cite{kulesza2012determinantal} has been already discussed and proved in multiple prior works.
We primarily focus on the main algorithms introduced in this paper: \textbf{QS-balanced} and the \textbf{MMR-balanced} algorithm.

\noindent
\textbf{Reducing Redundancy: }
A simple algorithm to reduce redundancy in the output summary is the following: let $R$ denote summary; at each step add the image $I \in S \setminus R$ which minimizes the following score
\begin{align*}
 \alpha \cdot A(q, I) - (1 - \alpha) \cdot \min_{J \in R} \sim(I,J), \eqlabel{1}
\end{align*}
where $\alpha \in [0,1]$. 
We use this expression, called the maximum marginal relevance, as a baseline in our experiments as well, and it is common in document summarization algorithm \cite{carbonell1998use, lin2011class}.

While this expression ensures the images are visibly diverse, it cannot focus on the features with respect to which diversity is desired by the user (as seen in Section~\ref{sec:results_gender}).
For example, it may ensure that the images in the summary have very different backgrounds, but cannot ensure the gender proportion of the people in the image summary is equal.
This leads us to using a diversity control set.

\noindent
\textbf{Diversity using a control set: }
To ensure visible diversity in the results we use a diversity control set $T_F$.

Adding the control set similarity score to expression \eqref{1}, we get the following relevance score for adding an image $I$ to a set $R$,
\begin{align*}
\textrm{mmod}_R(I) := (1-\alpha - \beta) \cdot A(q, I) + \alpha \cdot \min_{I_F \in T_F} \sim(I,I_F) -  \beta  \cdot \min_{J \in R} \sim(I,J), \eqlabel{2}
\end{align*}
where $\alpha, \beta \in [0,1]$.
The second term in the above expression now also aims to find the image in the control set $T_F$ most similar to $I$. 
If an image corresponding to $I_F$ has already been chosen, call it $J_{I_F}$, and $I$ has large similarity with $I_F$ as well, then we don't want to choose $I$. 
In this case values $\sim(I,I_F)$ and $\sim(I, J_{I_F})$ will be close and partially cancel each other, ensuring that the overall expression doesn't have the minimum value.

Recall that we use this scoring function as a baseline in our experiments as well. 
Furthermore, the expression \eqref{2} satisfies the diminishing-returns property.

\begin{lemma}[Submodularity of  \eqref{2}]
Let $f : 2^S \rightarrow \R$ be a function such that 
$f(R \cup \br{I}) - f(R) = - \textrm{mmod}_R(I).$
Then $f$ is submodular.
\end{lemma}
\noindent
\begin{proof}
For each $I$, $(\alpha \cdot A(q, I) + \beta \cdot \min_{I_F \in T_F} \sim(I,I_F))$ is constant and independent of the set $R$.
Consider two subsets $R_1 \subseteq R_2$. 
Then $$\min_{J \in R_1} \sim(I,J) \geq \min_{J \in R_2} \sim(I,J),$$ since the chance of an image in $R_2$ being similar to $I$ is larger than that for $R_1$.
Correspondingly, this score satisfies the diminishing-returns property.
\end{proof}

\noindent
\textbf{Alternate summarization relevance expression: }
Note that while the above algorithm can ensure diversity and non-redundancy, it has two major problems.

The first problem is that in the presence of a diversity control set, the primary aim is to ensure that diversity in the output set of images with respect to the features in the control set, and not the overall feature space.  
For such a task, the score $\min_{J \in R} \sim(I, J)$ may not ensure complete diversity with respect to the features of the diversity control set due to the additional goal of reducing redundancy.  
This was also observed in the empirical results presented in Section~\ref{sec:results_gender}; the standard deviation of the fraction of women in top results was higher for \textbf{MMR-balanced} results compared to \textbf{QS-balanced} results.
Hence we can try to slightly relax the goal of reducing redundancy to ensure better diversity with respect to diversity control set features.

The second problem is the time complexity. The iterative algorithm, based on choosing the image with the lowest score according to (2), is very slow. 
This is due to the fact that it has to evaluate the non-redundancy score $\min_{J \in R} \sim(I, J)$ at each step of the algorithm.
Once again, we can instead use $T_F$ directly to ensure diversity and reduce the time complexity.

This leads us to our main algorithm, which addresses both these issues.
Given the parameter $\alpha \in [0,1]$ and a query $q$, for each $I_F \in T_F$, our primary scoring function $\DS_{I_F}^q : S \rightarrow \R$ is the following:
\begin{align*}
\DS_{I_F}^q(I) =              \alpha \cdot \sim(I_F, I) + (1 - \alpha) \cdot A(q, I).
\end{align*}
We can show that this algorithm also corresponds to the diminishing returns property.
Furthermore, since it does not include any term to reduce redundancy by checking already chosen elements, using appropriate pre-processing (as mentioned in Section~\ref{sec:experiments}) it is much faster than \textbf{MMR-balanced}.

\noindent
\textbf{Diminishing-returns property of \textbf{QS-balanced} (Algorithm~\ref{algo:main}):} 
To show that  \textbf{QS-balanced} also satisfies the diminishing returns property, 
we will present an alternative iterative algorithm which outputs the same set as  \textbf{QS-balanced}  (Algorithm~\ref{algo:main}). 
For simplicity, assume that the size of the desired summary is a multiple of $|T_F|$.
Let $U : T_F \rightarrow 2^S$, be the following function
$U(I_F) := \br{I \in S \mid I_F = \arg\min_{I_F' \in T_F} \sim(I, I_F') }.$
Consider an iterative algorithm that adds one image to the final subset $R$ in each iteration. 
The image is chosen according to the following score function:
\begin{align*}
\DDS_R(I) := \begin{cases}
\frac{u}{2^n},  &\textrm{ if } \exists I_{F_1}, I_{F_2} \in T_F, \\ &\text{ s.t., } I_{F_1} \neq I_{F_2}, I = \arg \min_{I' \in U(I_{F_1}) \setminus R} \DS_{I_{F_1}}^q(I'), \\ &|U(I_{F_1}) \cap R| = n, \text{ and} \\ &|U(I_{F_2}) \cap R| > n \\
\frac{l}{2^n}, &\textrm{ otherwise,}
\end{cases}  \eqlabel{3}
\end{align*}
where $U(I_F)$ is as defined earlier and $u,l \in \R$ are numbers such that $l < u \leq 2l$.
Then we can prove the following theorems about the expression and its relation to (3).

\begin{theorem}
Given a dataset $S$, diversity control set $T_F$, query $q$, query relevance algorithm $A$ and numbers $u,l$, such that $l < u \leq 2l$, the set returned by Algorithm~\ref{algo:main} is the same as the set returned by the iterative algorithm using the scoring function \eqref{3}.
\end{theorem}
\noindent
\begin{proof}
As mentioned earlier (Figure~\ref{fig:model}), Algorithm~\ref{algo:main} is based on the constructing a $|T_F| \times |S|$ matrix using scores $\DS_{I_{F}}^q(I)$, and then sorting each row of the matrix. 
The images are finally chosen by taking images first from the first column, then second column and so on.
$\DDS$ score creates a similar ordering.

The first image chosen for any $I_F \in T_F$, since for all of them $|U(I_{F}) \cap R|  = 0$. The image chosen will be the image will have the best score with respect to $I_F$, i.e., $I = \arg \min_{I' \in U(I_{F_1})} \DS_{I_{F_1}}^q(I')$.
This corresponds to Step 9 of Algorithm~\ref{algo:main}.
Now $|U(I_{F}) \cap R|  = 1$ and for all other $I_F' \neq I_F$, $|U(I_{F}) \cap R|  = 0$, the iterative algorithm will next choose an image corresponding to a different $I_F' \neq I_F$ (since $u > l$), thus enforcing the loop in Step 8 of Algorithm~\ref{algo:main}.

Once one image is chosen for each $I_F$, the counter $n$ will increase and the same process will be repeated.
Since we assumed that the size of the chosen subset is a multiple of $|T_F|$, the ordering in which each $I_F$ is addressed does not matter.
\end{proof}
Note that the above expression can be modified for the case when the size of the desired summary is not a multiple of $|T_F|$.
To do so, one just has to fix an ordering for $I_{F_1}, I_{F_2}$ according to the scores $\br{\min_{I \in U(I_{F}) \setminus R} \DS_{I_{F}}^q(I)}_{I_F}$.

Having established the above equivalence, we can also show that the expression satisfies the diminishing-returns property.
\begin{lemma}[Submodularity of \eqref{3}]
Let $f : 2^S \rightarrow \R$ be a function such that 
$f(R \cup \br{I}) - f(R) = \DDS_R(I).$
Then $f$ is submodular.
\end{lemma}
The fact that the above function is submodular is in line with other functions considered for diverse image summarization, for example \cite{celis2018fair, tschiatschek2014learning}.
\begin{proof}
Consider two subsets $R_1 \subseteq R_2$. 
Let $n_1 := \lfloor |R_1| / |T_F| \rfloor $ and $n_2 := \lfloor |R_2| / |T_F| \rfloor $.
Assume that $I \in U(I_F)$. 
There are two cases that we need to address.

Case 1, $n_1 = n_2$. In this case, if $I = \arg \min_{I' \in U(I_{F}) \setminus R_2} \DS_{I_{F}}^q(I')$, then $\DDS_{R_1}(I) = \DDS_{R_2}(I) = u/2^{n_1}$.
If $I$ does not satisfy this condition and is not the image with the best score for $I_F$ in this iteration, then $\DDS_{R_1}(I) = \DDS_{R_2}(I) = l/2^{n_1}$.
For both cases, the score of $I$ is equal for $R_1$ and $R_2$.

Case 2, $n_1 < n_2$. In this case, there are two sub-cases. 
Either $I \neq \arg \min_{I' \in U(I_{F}) \setminus R_1} \DS_{I_{F}}^q(I')$ and  $I \neq \arg \min_{I' \in U(I_{F}) \setminus R_2} \DS_{I_{F}}^q(I')$. Then $\DDS_{R_1}(I) = l/2^{n_1}$ and $\DDS_{R_2}(I) = l/2^{n_2}$.
Since $n_1 < n_2$, we have that 
\[\frac{l}{2^{n_1}} > \frac{l}{2^{n_2}} \implies \DDS_{R_1}(I) \geq \DDS_{R_2}(I).\]
The other sub-case is that $I \neq \arg \min_{I' \in U(I_{F}) \setminus R_1} \DS_{I_{F}}^q(I')$ and  $I = \arg \min_{I' \in U(I_{F}) \setminus R_2} \DS_{I_{F}}^q(I')$.
Then $\DDS_{R_1}(I) = l/2^{n_1}$\\ and $\DDS_{R_2}(I) = u/2^{n_2}$.
Since $n_1 \leq n_2 - 1$ and $u < 2l$, we have that
\[\DDS_{R_1}(I) = \frac{l}{2^{n_1}} > \frac{u}{2^{n_1 + 1}} \geq \frac{u}{2^{n_2}} = \DDS_{R_2}(I).\]
Hence the score $\DDS$ follows the diminishing returns property for all cases.
\end{proof}

\section{Additional Empirical Results on Occupations Dataset}

In this section, we presents additional details and empirical results for Occupations dataset.
While the results in the main paper represent the best choice of parameters for the algorithm, such as $\alpha$ value or the diversity control set, we also present here the empirical results corresponding to varying parameters so as to motivate the choices made for the simulations in the main paper.

\subsection{Diversity Control Sets} \label{sec:occ_div_control_set}
\begin{figure}[t]
\centering     
\subfigure[Diversity Control Set - 1]{
\includegraphics[width=0.9\linewidth]{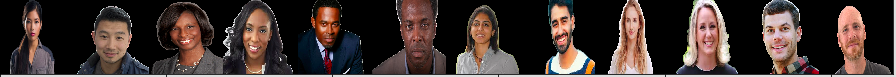}}
\subfigure[Diversity Control Set - 2]{
\includegraphics[width=0.9\linewidth]{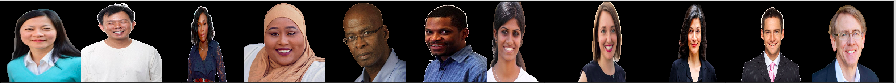}}
\subfigure[PPB Control Set - 1]{
\includegraphics[width=0.9\linewidth]{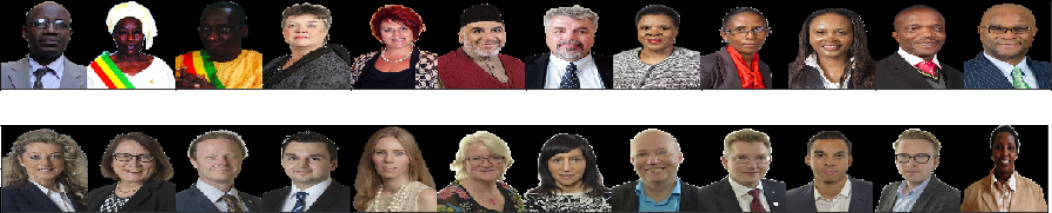}}
\subfigure[PPB Control Set - 2]{
\includegraphics[width=0.9\linewidth]{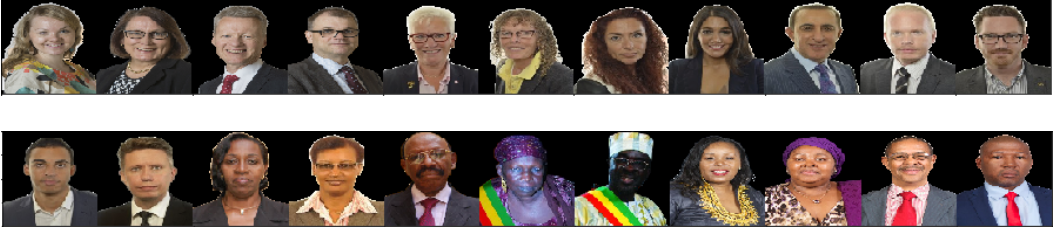}}
\caption{Occupations dataset: Diversity Control Sets used in the experiments.
The first two diversity controls (a) and (b) are hand-picked while the last two (c) and (d) were randomly sampled from the PPB dataset.
}
\label{fig:div_control_set}
\end{figure}

For Occupations dataset, we evaluate our approach on four different small (10-30 image) diversity control sets in order to evaluate the effect of the diversity control set on the end result. 
Two sets (Diversity Control Set-1 and Diversity Control Set-2) are hand selected by the authors using images from Google results, and are intended to be diverse with respect to presented gender and skin color.
The other two sets (PPB Control Set-1 and PPB Control Set-2) are generated by randomly sub-sampling from the Pilot Parliaments Benchmark Dataset  \cite{buolamwini2018gender}. This dataset has gender and skin-tone labelled images, and we select images uniformly at random conditioned on selecting equal number of men and women and equal number of people from all skin-tones.
The diversity control sets are presented in Figure~\ref{fig:div_control_set}.

\subsection{Intersectionality Results for all algorithms} \label{sec:intersec_occ}
We present the detailed intersectionality comparison with all baselines in the following table.
This is an extension of Table~\ref{tbl:int_div}.
The performance of \textbf{QS-balanced} algorithm can be observed to be better than other baselines in terms of intersectional diversity.

\begin{table}[H]
\centering
\small
\begin{tabular}{ |c |c |c |c | c| }
\hline 
Algorithm & \makecell{\% gender\\ stereotypical\\ with fair skin} & \makecell{\% gender \\anti-stereotypical\\ with fair skin} & \makecell{\% gender\\ stereotypical\\ with dark skin} & \makecell{\% gender \\anti-stereotypical\\ with dark skin}  \\
\hline
\hline
QS-balanced & 0.46 (0.14) &  0.37 (0.14) & 0.09 (0.05) & 0.08 (0.05)\\
MMR-balanced  & 0.46 (0.17)  & 0.39 (0.18) & 0.09 (0.06)  & 0.06 (0.04) \\
\hdashline
Google  & 0.60 (0.20) & 0.24 (0.21) & 0.11 (0.08) & 0.05 (0.07)\\
\hdashline
MMR  & 0.57 (0.21) & 0.30 (0.21) & 0.07 (0.06) & 0.05 (0.05)\\
DET  & 0.52 (0.12) &  0.33 (0.12) & 0.09 (0.05) &  0.06 (0.05) \\
AUTOLABEL  & 0.54 (0.16) & 0.31 (0.16) & 0.09 (0.06) & 0.06 (0.04) \\
AUTOLABEL-RWD  & 0.56 (0.19) & 0.30 (0.19) & 0.08 (0.06) & 0.05 (0.05)\\
\hline
\end{tabular}
\caption{Occupations dataset: Intersectionality comparison with all baselines.}
\label{tbl:int_div_all}
\end{table}

\subsection{Results for different diversity control sets} \label{sec:diff_control_sets_occ}
\begin{figure}[H]
\centering     
\subfigure[\scriptsize{Gender diversity comparison}]{
\includegraphics[width=0.48\linewidth]{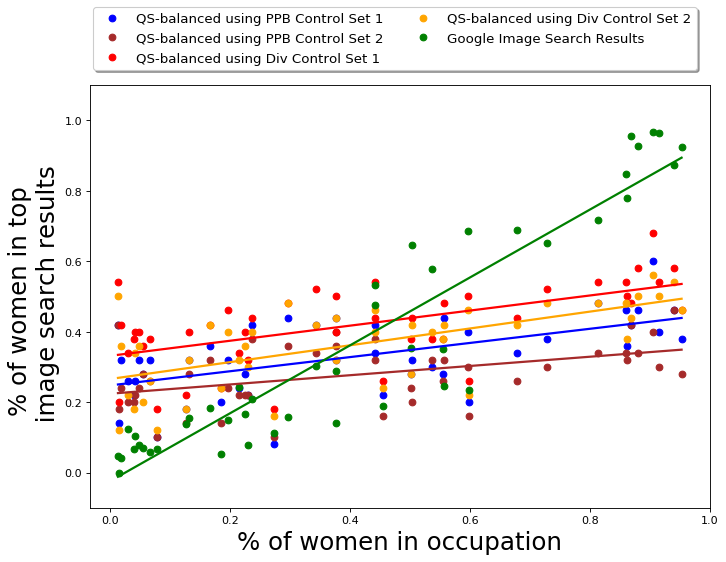}}
\subfigure[\scriptsize{Skin-tone diversity comparison}]{
\includegraphics[width=0.48\linewidth]{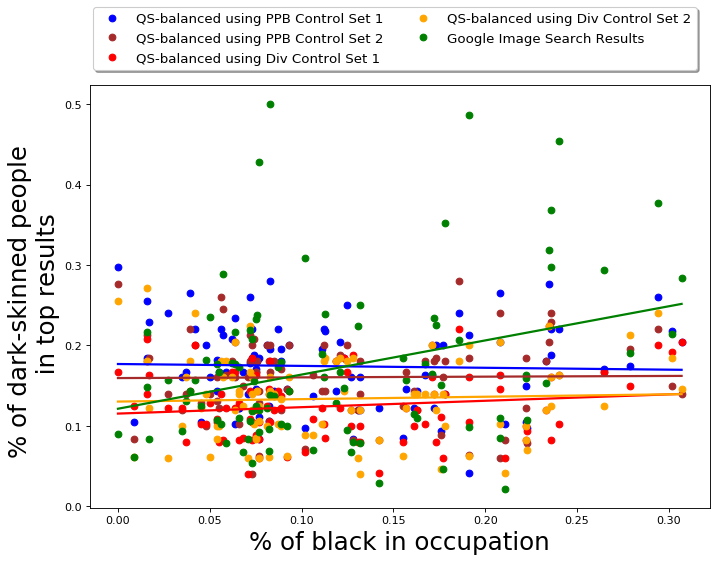}}
\caption{Occupations dataset: Gender and skin-tone diversity comparison of results of \textbf{QS-balanced} algorithm on different diversity control sets.
For gender, using any of the diversity control sets results in a more gender-balanced output.
For skin-tone, using PPB Control Set-1 results in the best results amongst all diversity control sets. For most occupations, the top Google images are have much larger or much smaller fraction of images of dark-skinned people.
}
\label{fig:occ_comparison_diff_divs}
\end{figure}

As noted earlier, we use 4 different diversity control sets in our empirical evaluations.
The results presented in the paper correspond to evaluation using PPB-control set 1.
We provide the diversity comparison for different diversity control sets in Figure~\ref{fig:occ_comparison_diff_divs}.

\subsection{Results for different compositions of diversity control sets} \label{sec:diff_comp_occ}

\begin{figure}[H]
\centering     
\subfigure[\scriptsize{Different fraction of women in Diversity Control Set}]{
\includegraphics[width=0.47\linewidth]{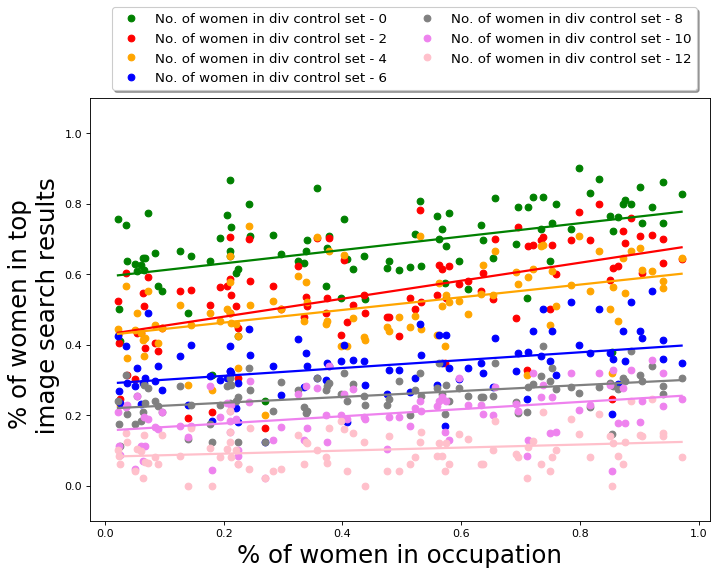}}
\subfigure[\scriptsize{Different fraction of dark-skinned in Diversity Control Set}]{
\includegraphics[width=0.47\linewidth]{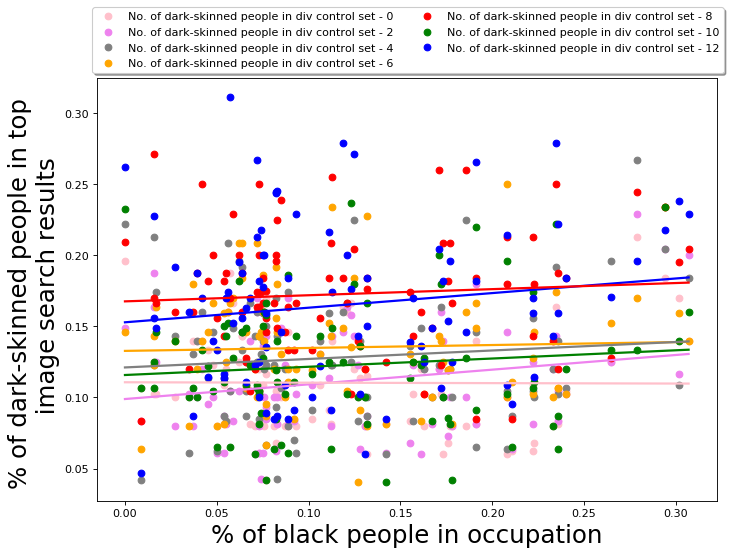}}
\subfigure[\scriptsize{Different fraction of dark-skinned women in Diversity Control Set}]{
\includegraphics[width=0.5\linewidth]{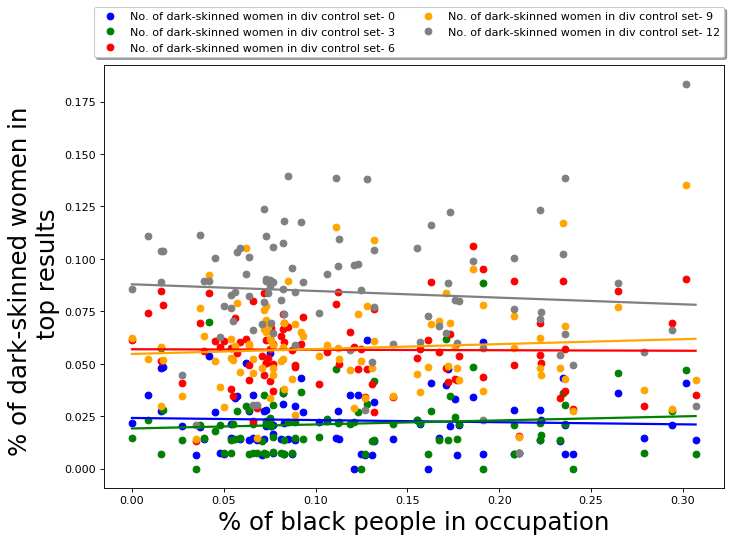}}
\caption{Occupations dataset: Performance \textbf{QS-balanced} algorithm on diversity control sets with different compositions.
}
\label{fig:occ_comparison_diff_divs_2}
\end{figure}

To explicitly see the impact of diversity control on the diversity of the output of the algorithm, we can vary the content of the diversity control set and observe the corresponding changes in the results.
We first vary the fraction of women in the diversity control set. 
The diversity control sets are randomly chosen for the PPB-dataset, while maintaining the desired gender ratio.
The results for different diversity control sets are presented in Figure~\ref{fig:occ_comparison_diff_divs_2}a. The figure shows that increasing the fraction of women in the diversity control set leads to an increase in the fraction of women in the output set.

Similarly, increasing the fraction of images of dark-skinned people in the diversity control set leads to an increase in fraction of images of dark-skinned people in the output; this is shown in Figure~\ref{fig:occ_comparison_diff_divs_2}b.
Finally, Figure~\ref{fig:occ_comparison_diff_divs_2}c shows the impact of variation of images of dark-skinned women in the control set on the output.
While the fraction of dark-skinned women still increases, it seems to be upper bounded by the fraction of images of dark-skinned women in the dataset.

\subsection{Results for different $\alpha$ values} \label{sec:diff_alpha_occ}
\begin{figure}[H]
\centering     
\subfigure[\scriptsize{QS-balanced}]{
\includegraphics[width=0.48\linewidth]{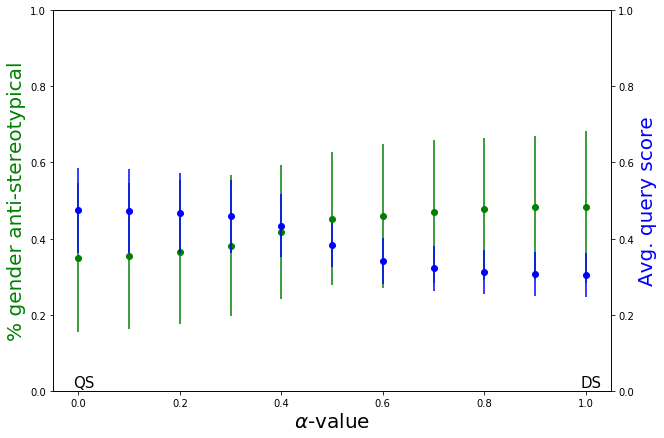}}
\subfigure[\scriptsize{MMR-balanced}]{
\includegraphics[width=0.48\linewidth]{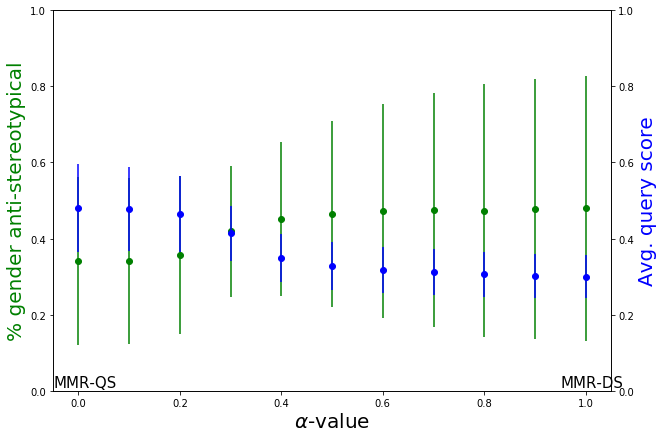}}
\caption{Occupations dataset: Gender  diversity and query similarity comparison of results of \textbf{QS-balanced}  and \textbf{MMR-balanced} algorithms for different $\alpha$-values.
}
\label{fig:occ_comparison_diff_alphas}
\end{figure}

\begin{figure}[H]
\centering     
\subfigure[\scriptsize{QS-balanced}]{
\includegraphics[width=0.48\linewidth]{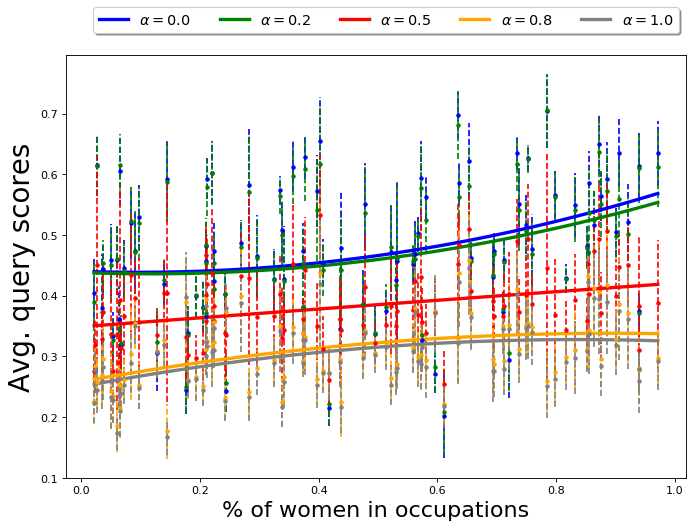}}
\subfigure[\scriptsize{MMR-balanced}]{
\includegraphics[width=0.48\linewidth]{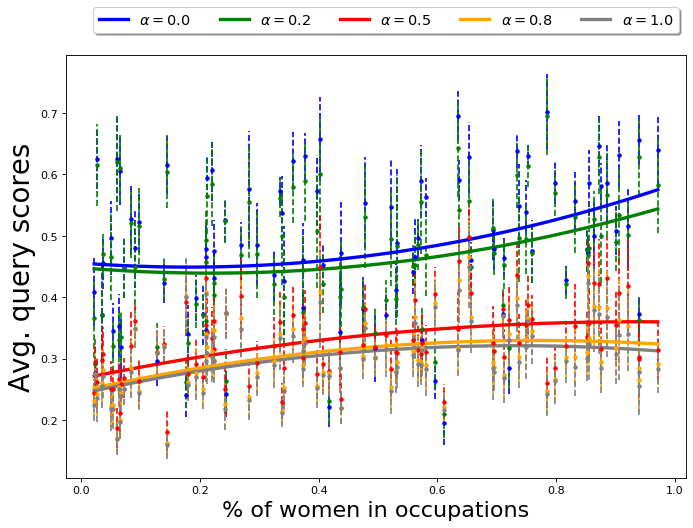}}
\caption{Occupations dataset: Accuracy of results of \textbf{QS-balanced}  and \textbf{MMR-balanced} algorithms for different $\alpha$-values.
}
\label{fig:occ_comparison_diff_alphas_2}
\end{figure}

We vary the quality-fairness parameter $\alpha$ and look at its impact on the performance of our algorithms.
The diversity results are presented in Figure~\ref{fig:occ_comparison_diff_alphas}, while Figure~\ref{fig:occ_comparison_diff_alphas_2} expands on the accuracy for different alphas.

For both \textbf{QS-balanced} and \textbf{MMR-balanced}, the fraction of gender anti-stereotypical images increases as the $\alpha$ value increases.
With an increase in fairness, a loss in accuracy is expected.
While the figure shows a small change in average query scores, the standard deviation of the scores seem to be decreasing as well, showing that as $\alpha$ increases, the dependence on the query decreases.
Hence a balance between query similarity and diversity score has to be maintained by choosing an appropriate value of $\alpha$, such as 0.5.

\subsection{Results for different summary sizes} \label{sec:occ_diff_sizes}
{
While the results we have presented so have been with respect to a summary of size 50. 
However, the size of the summary can depend on the application and the results in the first page of any web-search application will depend on the size of the screen or the device being used.
Correspondingly, it is important to analyze the results for different summary sizes as well.
}

\begin{figure}[t]
\centering     
\subfigure[\scriptsize{Fraction of images of gender anti-stereotypical people vs summary size}]{
\includegraphics[width=0.48\linewidth]{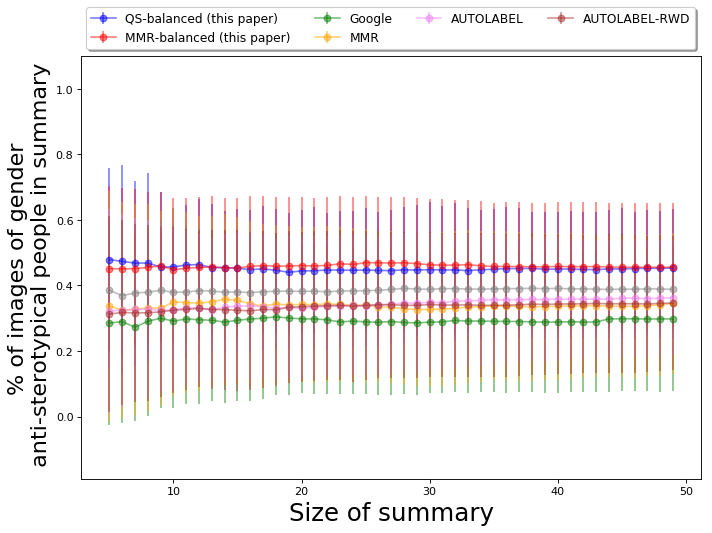}}
\subfigure[\scriptsize{Fraction of images of dark-skinned people vs summary size}]{
\includegraphics[width=0.48\linewidth]{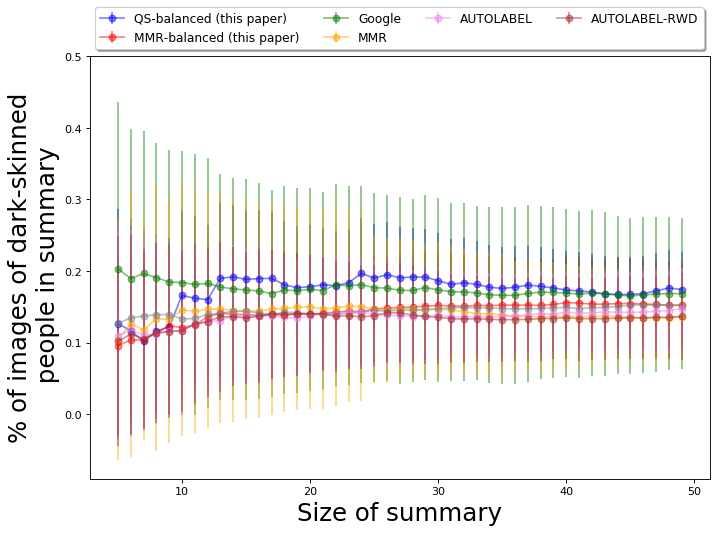}}
\caption{Occupations dataset: Variation of fraction of anti-stereotypical images vs size of summary for all algorithms.
}
\label{fig:occ_comparison_summary_size}
\end{figure}

\begin{figure}[t]
\centering     
\subfigure[\scriptsize{Fraction of images of gender anti-stereotypical people vs ground truth}]{
\includegraphics[width=0.48\linewidth]{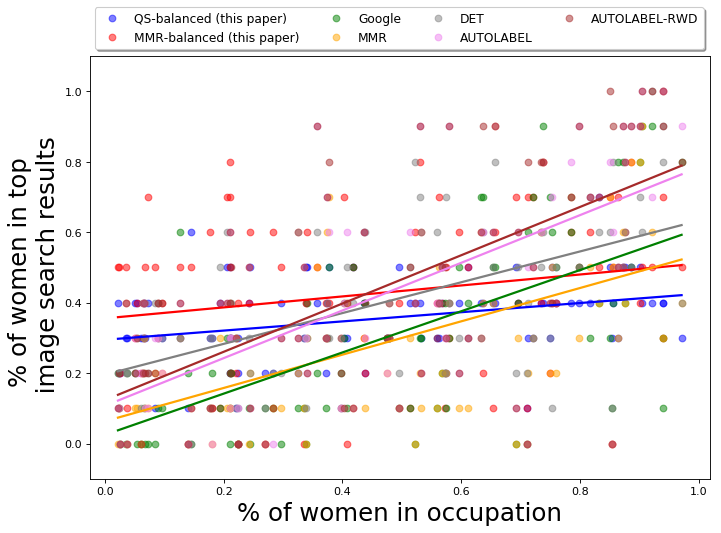}}
\subfigure[\scriptsize{Fraction of images of dark-skinned people vs ground truth}]{
\includegraphics[width=0.48\linewidth]{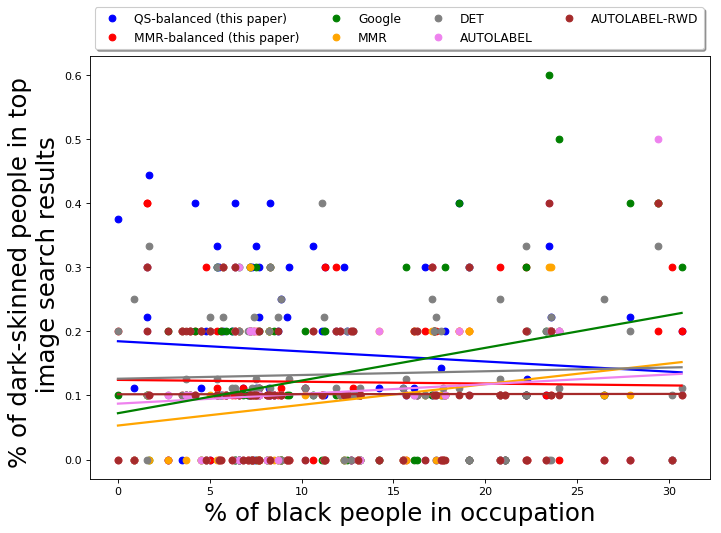}}
\caption{Occupations dataset: Fraction of anti-stereotypical images for summary size 10.
}
\label{fig:occ_comparison_10}
\end{figure}

\begin{figure}[t]
\centering     
\subfigure[\scriptsize{Fraction of images of gender anti-stereotypical people vs ground truth}]{
\includegraphics[width=0.48\linewidth]{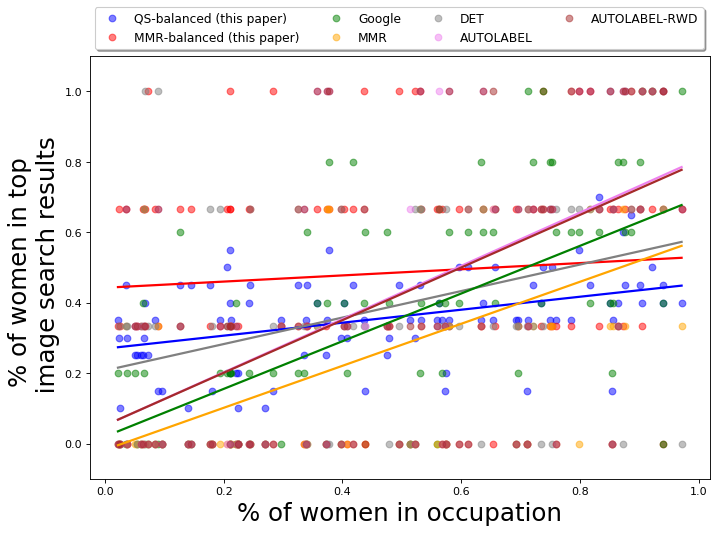}}
\subfigure[\scriptsize{Fraction of images of dark-skinned people vs ground truth}]{
\includegraphics[width=0.48\linewidth]{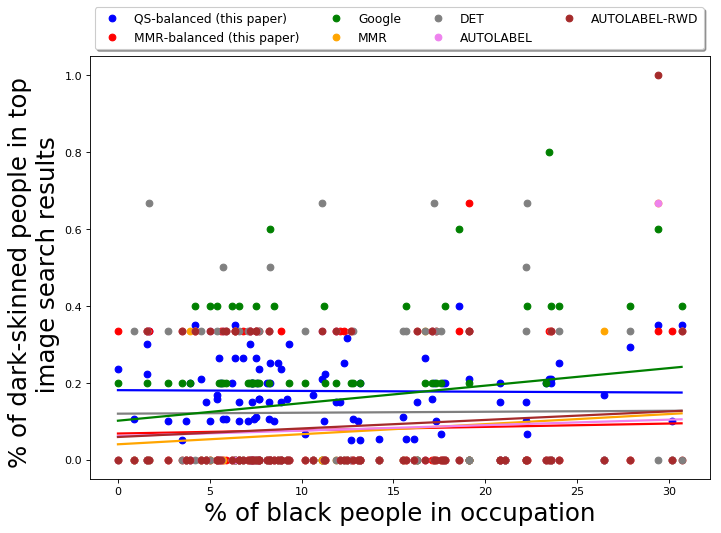}}
\caption{Occupations dataset: Fraction of anti-stereotypical images for summary size 20.
}
\label{fig:occ_comparison_20}
\end{figure}

{
For \textbf{QS-balanced}, \textbf{MMR-balanced} and the baselines, we look at the average fraction of images of gender anti-stereotypical and dark-skinned people in the top $k$ results, where $k$ ranges from 2 to 50; the average is taken over all occupations.
The results are presented in Figure~\ref{fig:occ_comparison_summary_size}.
We also present the gender and skintone-diversity comparison of our method vs baselines for summary sizes 10 and 20 in Figure~\ref{fig:occ_comparison_10} and  Figure~\ref{fig:occ_comparison_20}.
}

{
The figures shows that \textbf{QS-balanced} and \textbf{MMR-balanced} return a larger fraction of gender anti-stereotypical images for all summary sizes. 
With respect to skin-type, Google results seem to have a larger value for average fraction of dark-skinned people for smaller summary sizes; however, the performance of \textbf{QS-balanced} in this respect is similar to better for larger summary sizes.
Furthermore, Google results also have a significantly larger standard deviation, implying that the fraction of dark-skinned people is also much lower than average for some occupations.
}

\subsection{Similarity and Non-redundancy comparison for Occupations dataset} \label{sec:non_red_appendix}

\begin{figure}[t]
\centering     
\subfigure[Avg. query scores - Comparison with baselines]{
\includegraphics[width=0.48\linewidth]{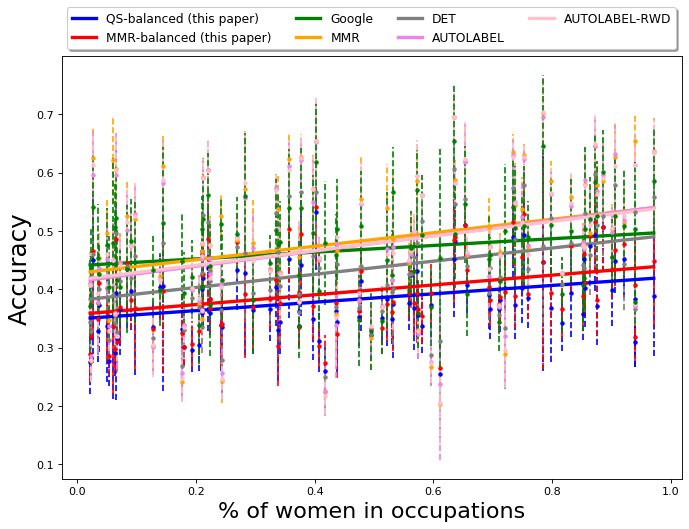}}
\subfigure[Non-redundancy scores - Comparison with baselines]{
\includegraphics[width=0.48\linewidth]{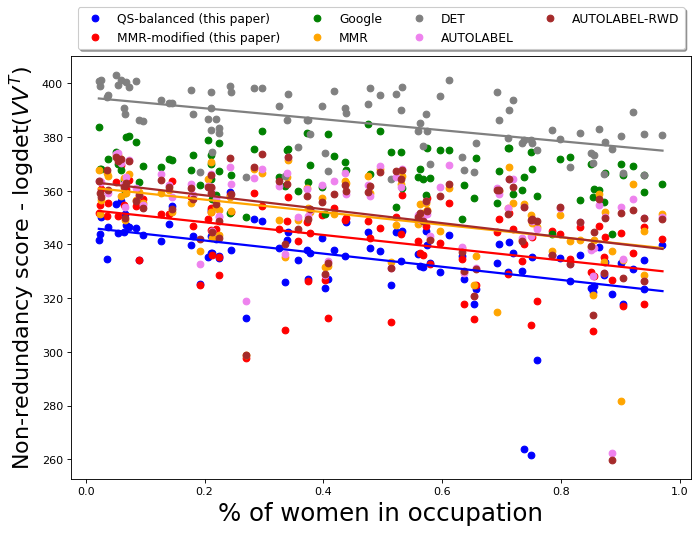}}
\caption{Occupations dataset: (a) Comparison of accuracy, as measured using mean query similarity scores, of top 50 results across all occupations. For each occupation, we also plot mean similarity to the query control set and the standard deviation using the dotted lines. The mean similarity score of results of all algorithms are close to each other, showing that using diversity control set does not adversely impact the accuracy.
(b) Comparison of non-redundancy scores. As expected, the results from \textbf{DET} have the largest non-redundancy score, measured as the log of determinant of the product of feature matrix the output images and its transpose. 
The non-redundancy scores of \textbf{QS-balanced} and \textbf{MMR-balanced} are the lowest, perhaps due to enforcing fairness constraints using the diversity control set. 
}
\label{fig:comparison_sim_scores}
\end{figure}

As mentioned earlier, the accuracy for Occupations dataset is measured using average query similarity, i.e., similarity to the set of images corresponding to the given query.
We present the accuracy comparison for our methods and baselines in Figure~\ref{fig:comparison_sim_scores}.
The accuracy score of results of all algorithms are close to each other, showing that using diversity control set does not adversely impact the accuracy.

The second figure also presents the non-redundancy comparison of our methods and baselines.
The non-redundancy measure used is the log of the determinant of the feature kernel matrix, i.e., if for a summary $S$, if $V_S$ is the matrix with columns representing the feature vectors of the images in $S$, then the non-redundancy is measured as $\log \det (V_SV_S^\top)$ (the determinant can be pretty large and computationally more difficult to calculate, hence the logarithm).
As expected, the results from \textbf{DET} have the largest non-redundancy score.
The non-redundancy scores of \textbf{QS-balanced} and \textbf{MMR-balanced} are the lowest, perhaps due to enforcing fairness constraints using the diversity control set. 
However, as we saw earlier, non-redundancy does not imply diversity with respect to the socially salient attributes.

\subsection{Occupation accuracy of QS-balanced algorithm} \label{sec:occ_acc_alt}
\begin{figure}[H]
\centering     
\includegraphics[width=\linewidth]{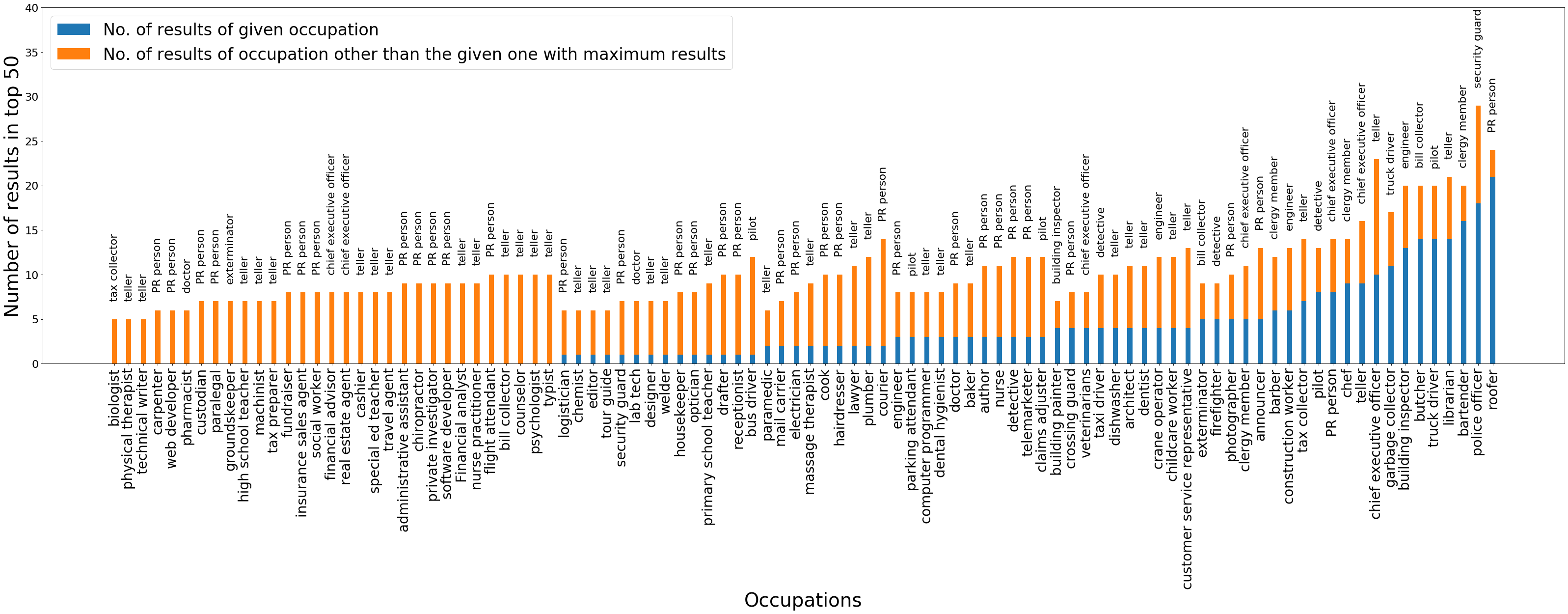}
\caption{Occupations dataset: Accuracy comparison of results of \textbf{QS-balanced} algorithm for different occupations.
For each occupation and its summary, we present the number of images belonging to that occupation in the summary, as well as the other occupation with highest number of images in the summary.
}
\label{fig:acc_bar_graph}
\end{figure}

Finally, we also present the accuracy of the results of QS-balanced algorithm.
The accuracy is measured as the number of images in the summary belonging to the queried occupation.
The results for this accuracy are presented in Figure~\ref{fig:acc_bar_graph}.
Note that accuracy is not a good measure of quality in this case; this is because a lot of occupations have similar looking images. For example, images of lawyers and financial analysts are very similar, images of doctors and pharmacists are very similar.
Hence when using image similarity as a method of query matching, one cannot expect the matched images to always belong to the same query.
This problem is relatively less visible for CelebA dataset, since in that case the query similarity algorithm is more specialized to the dataset.

We also present the bar graph for when 1-norm is used, instead of cosine distance for similarity.
In this case, the accuracy is much worse and this is reason for using cosine-distance over 1-norm distance for all our simulations. 
\begin{figure}[H]
\centering     
\includegraphics[width=\linewidth]{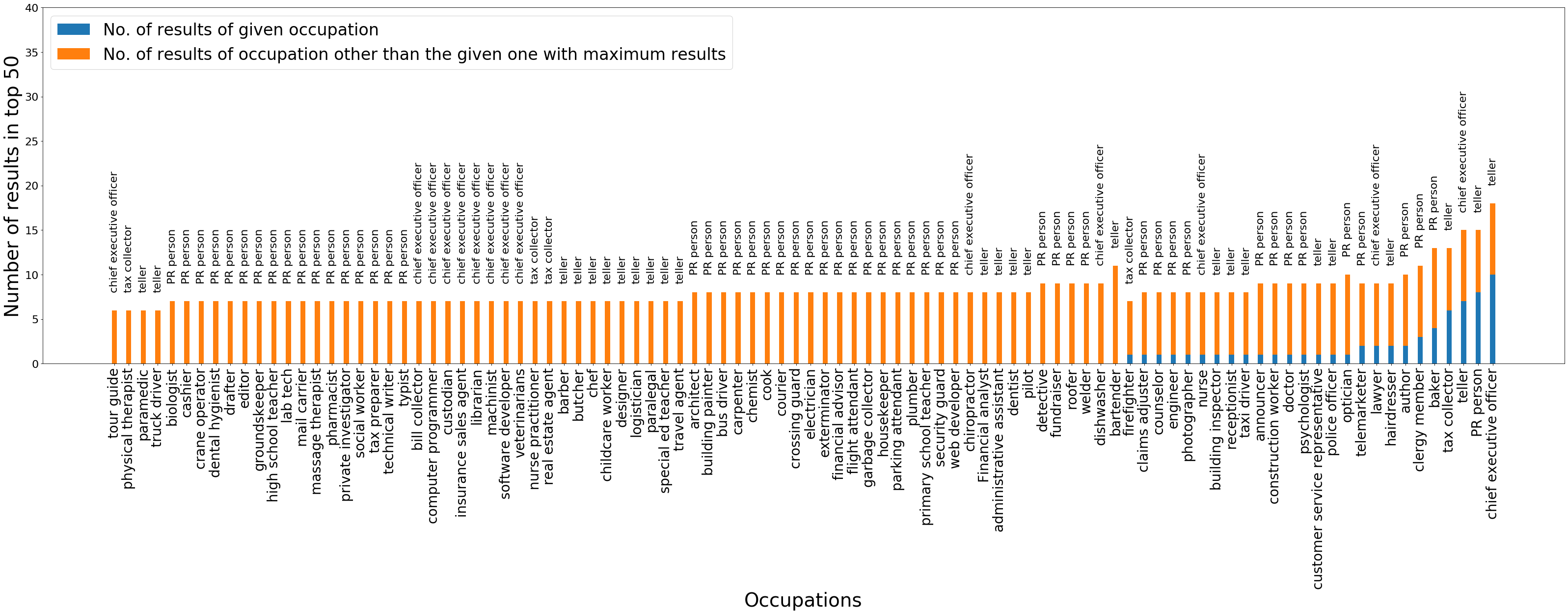}
\caption{Occupations dataset: Accuracy comparison of results of \textbf{QS-balanced} algorithm for different occupations using 1-norm for similarity.
}
\label{fig:accuracy_1_norm}
\end{figure}


\section{Additional Results on CelebA Dataset}\label{sec:celeba_add}

In this section, we present additional details and empirical results for CelebA dataset.
The additional results correspond to varying different parameters in the algorithm, such as $\alpha$ value or the diversity control set.

\subsection{Attributes of the dataset} \label{sec:celeba_attr}
{
We first present the list of facial attributes in the dataset and the fraction of images with a given attribute that are also labelled ``Female''.
}

\vspace{0.2in}
\begin{tabular}{|p{3cm}|p{3cm}||p{3cm}|p{3cm}|}
\hline
Attribute & Fraction of images of women with given attribute & Attribute & Fraction of images of women with given attribute \\
\hline
Heavy Makeup & 1.0 & Wearing Lipstick & 0.99 \\
Rosy Cheeks & 0.98 & Wearing Earrings & 0.96 \\
Blond Hair & 0.94 & Wearing Necklace & 0.94 \\
Arched Eyebrows & 0.92 & Wavy Hair & 0.82 \\
Attractive & 0.77 & Bangs & 0.77 \\
Pale Skin & 0.76 & Pointy Nose & 0.76 \\
Big Lips & 0.73 & High Cheekbones & 0.72 \\
No Beard & 0.7 & Brown Hair & 0.69 \\
Oval Face & 0.68 & Smiling & 0.65 \\
Mouth Slightly Open & 0.63 & Narrow Eyes & 0.56 \\
Blurry & 0.53 & Straight Hair & 0.52 \\
Black Hair & 0.48 & Receding Hairline & 0.39 \\
Wearing Hat & 0.3 & Bags Under Eyes & 0.29 \\
Bushy Eyebrows & 0.28 & Big Nose & 0.25 \\
Eyeglasses & 0.21 & Gray Hair & 0.15 \\
Chubby & 0.12 & Double Chin & 0.12 \\
5 o Clock Shadow & 0.0 & Bald & 0.0 \\
Goatee & 0.0 & Male & 0.0 \\
Mustache & 0.0 & Sideburns & 0.0 \\
\hline
\end{tabular}

\subsection{Diversity Control Sets} \label{sec:celeba_div_control_set}
\begin{figure}[t]
\centering     
\subfigure[Diversity Control Set - 1]{
\includegraphics[width=0.7\linewidth]{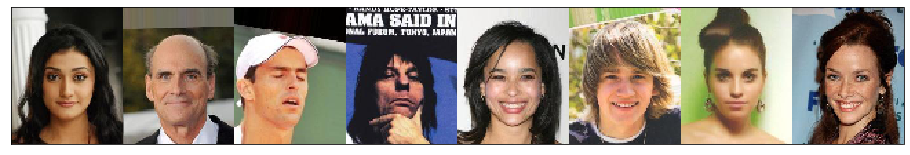}}
\subfigure[Diversity Control Set - 2]{
\includegraphics[width=0.7\linewidth]{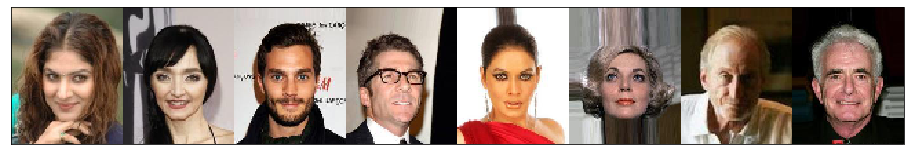}}
\subfigure[Diversity Control Set - 3]{
\includegraphics[width=0.7\linewidth]{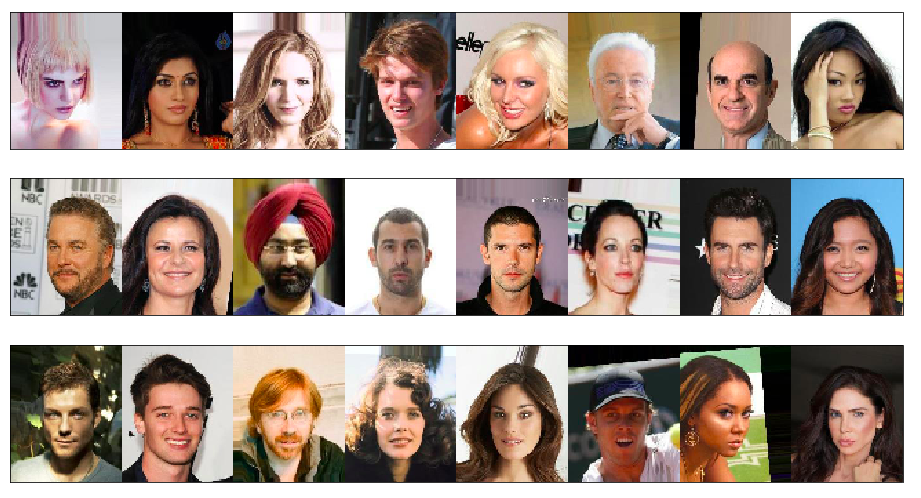}}
\subfigure[Diversity Control Set - 4]{
\includegraphics[width=0.7\linewidth]{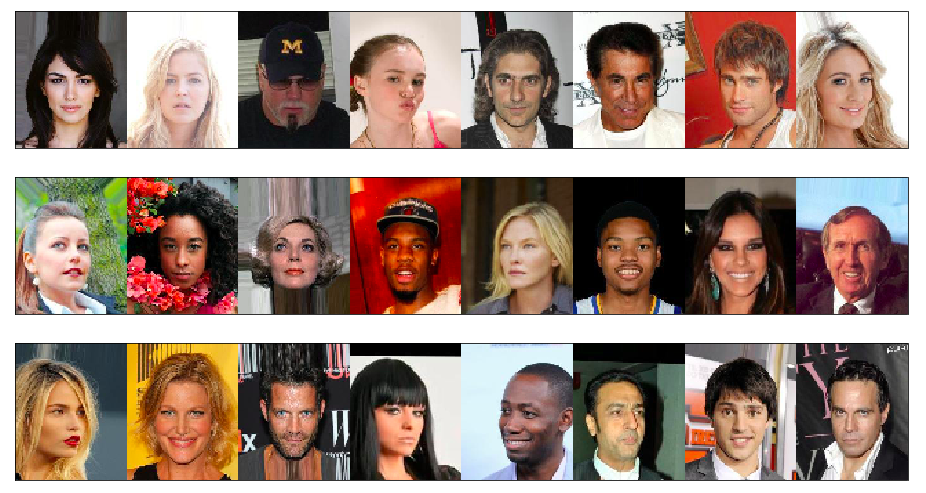}}
\caption{CelebA dataset: Diversity Control Sets used in the for empirical evaluation on CelebA dataset.
}
\label{fig:div_control_set_2}
\end{figure}

Once again, we will use four different diversity control sets for our evaluation, two of them have 8 images and the other two have 24 images; the exact images are provided in Section~\ref{sec:celeba_div_control_set}.
The diversity control sets are constructed by randomly sampling equal number of images with and without the ``Male'' attribute from the train set.
The diversity control sets are presented in Figure~\ref{fig:div_control_set_2}.

\subsection{Results by features}
\begin{figure}[t]
\centering     
\subfigure[\scriptsize{Gender diversity comparison}]{
\includegraphics[width=0.48\linewidth]{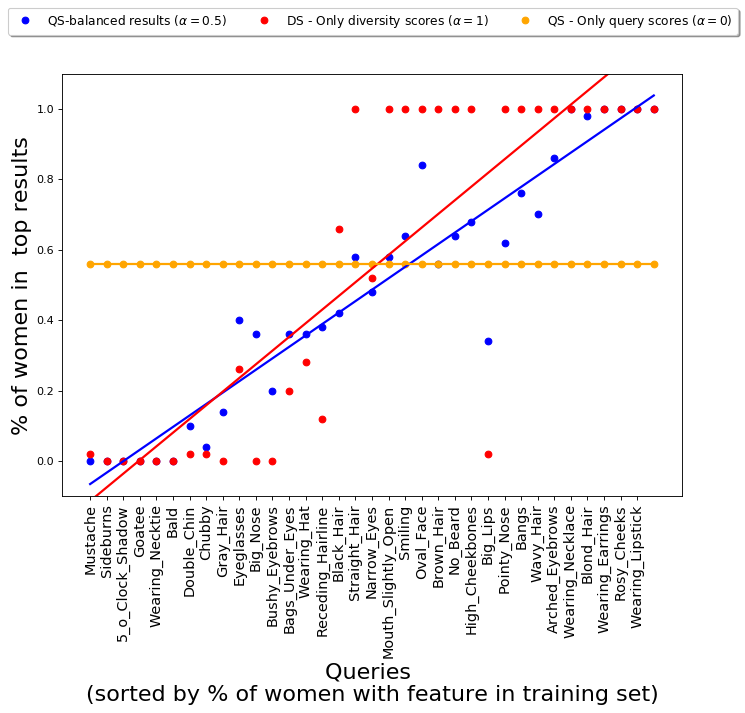}}
\subfigure[\scriptsize{Accuracy comparison}]{
\includegraphics[width=0.48\linewidth]{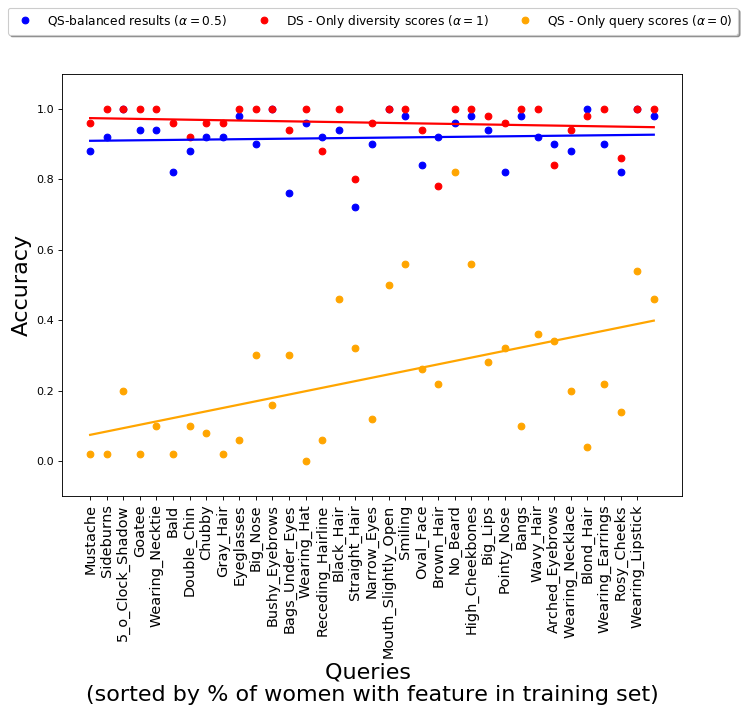}}
\caption{CelebA dataset: Gender and accuracy comparison of results of \textbf{QS-balanced} algorithm for all queries.
}
\label{fig:celeba_comparison_all}
\end{figure}

We first present the exact gender and accuracy results by features in Figure~\ref{fig:celeba_comparison_all}. 

\subsection{Results for different diversity control sets}
\begin{figure}[t]
\centering     
\subfigure[\scriptsize{Gender diversity comparison}]{
\includegraphics[width=0.48\linewidth]{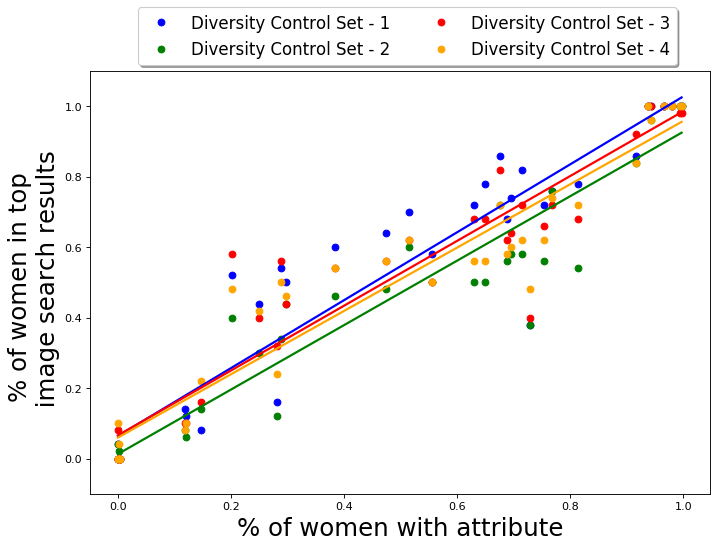}}
\subfigure[\scriptsize{Accuracy comparison}]{
\includegraphics[width=0.48\linewidth]{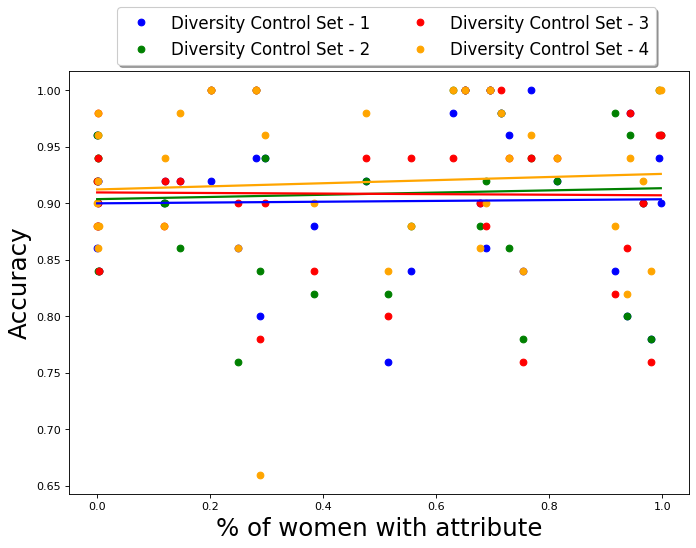}}
\caption{CelebA dataset: Gender diversity and accuracy comparison of results of \textbf{QS-balanced} algorithm on different diversity control sets.
For all the control sets, the performance with respect to gender diversity and accuracy seems to be similar.
}
\label{fig:celeba_comparison_diff_divs}
\end{figure}

As noted earlier, we use 4 different diversity control sets in our empirical evaluations.
The results presented in the paper correspond to evaluation using Diversity Control Set-4.
We provide the accuracy and diversity comparison for different diversity control sets in Figure~\ref{fig:celeba_comparison_diff_divs}.

\subsection{Results for different compositions of diversity control sets}

\begin{figure}[t]
\centering     
\subfigure[\scriptsize{Gender comparison}]{
\includegraphics[width=0.48\linewidth]{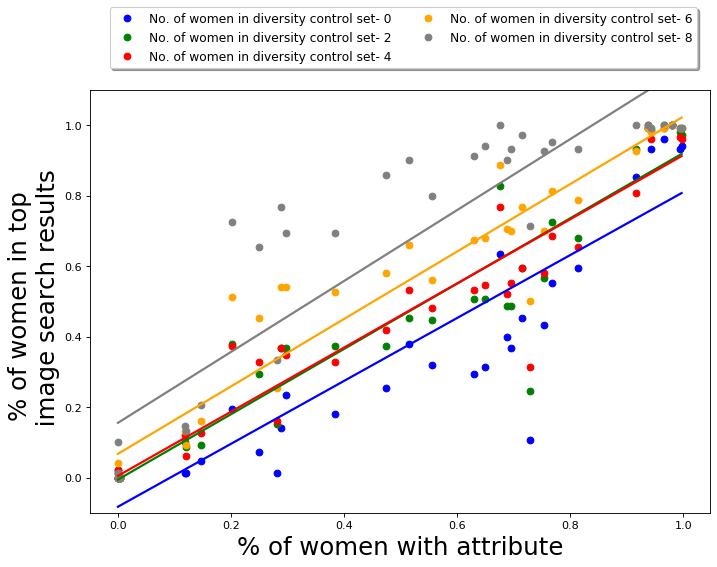}}
\subfigure[\scriptsize{Accuracy comparison}]{
\includegraphics[width=0.48\linewidth]{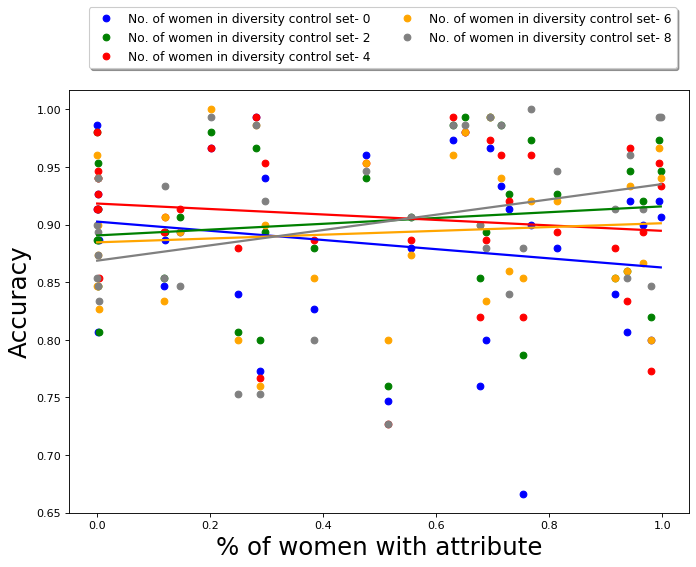}}
\caption{CelebA dataset: Performance of \textbf{QS-balanced} algorithm on diversity control sets with different compositions.
}
\label{fig:celeba_comparison_diff_divs_2}
\end{figure}

To explicitly see the impact of diversity control on the diversity of the output of the algorithm, we once again vary the content of the diversity control set and observe the corresponding changes in the results.
In this case, we only vary the fraction of women in the diversity control set. 
The diversity control sets are randomly chosen from the training dataset, while maintaining the desired gender ratio.
The results for different diversity control sets are presented in Figure~\ref{fig:celeba_comparison_diff_divs_2}. The figure shows that increasing the fraction of women in the diversity control set leads to an increase in the fraction of women in the output set.

\begin{figure}[t]
\centering     
\includegraphics[width=0.5\linewidth]{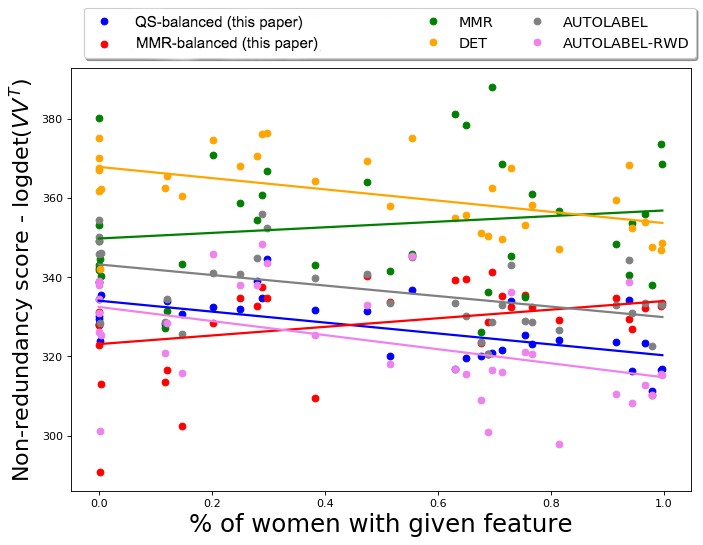}
\caption{CelebA dataset: Non-redundancy comparison of our methods vs baselines.
}
\label{fig:celeba_comparison_nonred}
\end{figure}

\subsection{Non-redundancy comparison} \label{sec:non_red_celeba}
{
Figure~\ref{fig:celeba_comparison_nonred} presents the non-redundancy comparison of our methods and baselines.
Recall that the non-redundancy measure used is the log of the determinant of the feature kernel matrix, i.e., if for a summary $S$, if $V_S$ is the matrix with columns representing the feature vectors of the images in $S$, then the non-redundancy is measured as $\log \det (V_SV_S^\top)$.
As expected, the results from \textbf{DET} have the largest non-redundancy score for most attributes.
However, once again, non-redundancy does not imply diversity with respect to the socially salient attributes.
}

\subsection{Results for different $\alpha$ values} \label{sec:diff_alpha_celeba}
\begin{figure}[t]
\centering     
\subfigure[\scriptsize{QS-balanced}]{
\includegraphics[width=0.48\linewidth]{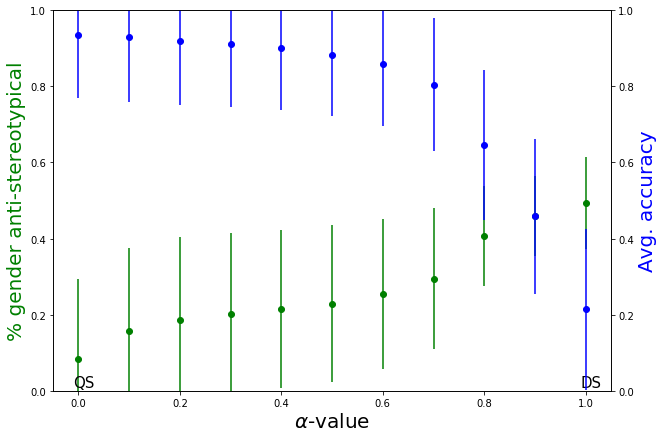}}
\subfigure[\scriptsize{MMR-balanced}]{
\includegraphics[width=0.48\linewidth]{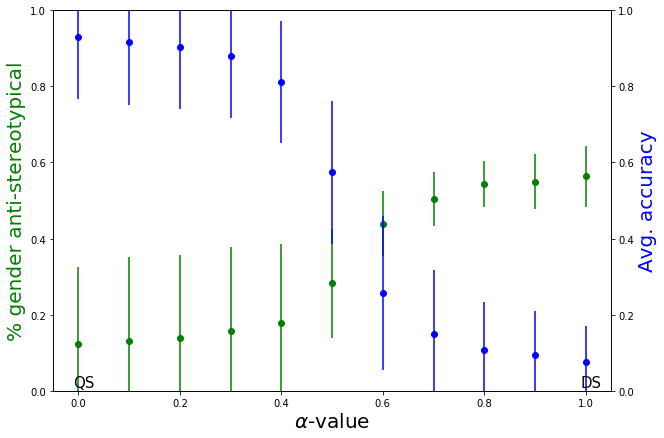}}
\caption{CelebA dataset: Gender  diversity and query similarity comparison of results of \textbf{QS-balanced}  and \textbf{MMR-balanced} algorithms for different $\alpha$-values.
}
\label{fig:celeba_comparison_diff_alphas}
\end{figure}

We vary the quality-fairness parameter $\alpha$ and look at its impact on the performance of our algorithms.
The results are presented in Figure~\ref{fig:celeba_comparison_diff_alphas}.

For both \textbf{QS-balanced} and \textbf{MMR-balanced}, the fraction of gender anti-stereotypical images increases as the $\alpha$ value increases.
However, increasing the $\alpha$ value results in a corresponding decrease in the accuracy, which is much more significant for \textbf{MMR-balanced} results.

\subsection{Results for different summary sizes} \label{sec:celeba_diff_sizes}
\begin{figure}[t]
\centering     
\subfigure[\scriptsize{Fraction of images of gender anti-stereotypical people vs summary size}]{
\includegraphics[width=0.48\linewidth]{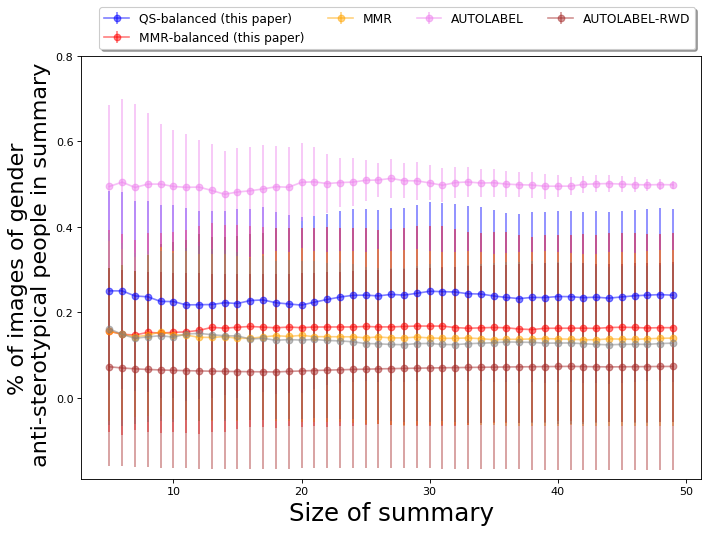}}
\subfigure[\scriptsize{Accuracy vs summary size}]{
\includegraphics[width=0.48\linewidth]{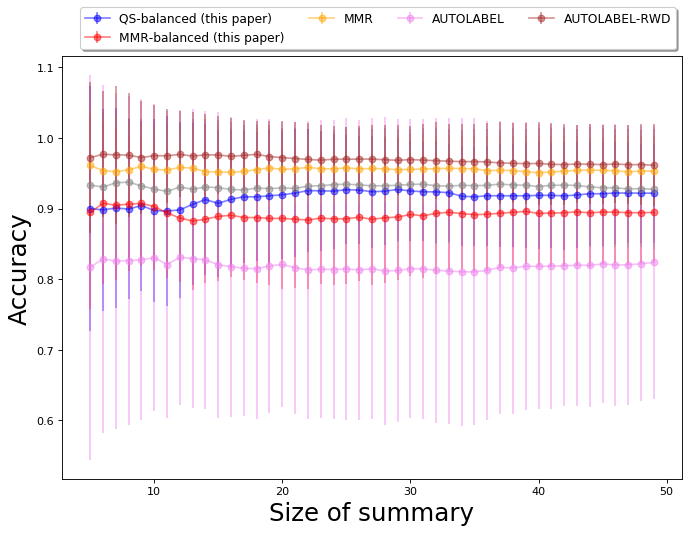}}
\caption{CelebA dataset: Variation of fraction of gender anti-stereotypical images and accuracy vs size of summary for all algorithms.
}
\label{fig:celeba_comparison_summary_size}
\end{figure}

{
Once again, we provide the performance of our algorithms and baselines for different summary sizes.
For \textbf{QS-balanced}, \textbf{MMR-balanced} and the baselines, we look at the average fraction of images of gender anti-stereotypical and dark-skinned people in the top $k$ results, where $k$ ranges from 2 to 50; the average is taken over all occupations.
The results are presented in Figure~\ref{fig:celeba_comparison_summary_size}.
The figures shows that \textbf{QS-balanced} returns a larger fraction of gender anti-stereotypical images for all summary sizes, compared to all baselines, other than \textbf{AUTOLABEL}. 
While \textbf{AUTOLABEL} is able to achieve better gender diversity, in this case, due to the good performance of the auto-gender classifier, simply using the partitions has an impact on accuracy of the summaries generated by \textbf{AUTOLABEL}.
The second figure shows that the accuracy of \textbf{AUTOLABEL} is the worst amongst all algorithms.
}

\end{document}